# Limits to rover miniaturisation and their implications for solar system exploration

Dr Stephen Edwards, Visiting Researcher, Isotope Group, School of Earth and Environmental Sciences, University of Manchester.

## Abstract

Modern planetary exploration is tightly mass constrained, and as a result, only one or two, 100 kg scale missions per target are typically undertaken. If smaller vehicles could be used, many more locales on a planetary target could be investigated, and due to redundancy through numbers, higher risk, scientifically interesting terrains explored.

Scaling issues are therefore examined for mobile semiautonomous rovers for solid body surface exploration throughout the solar system. Communications to a relay orbiter emerges as a major size constraint, due to power requirements and the $D^2$ increase in antenna gain with increasing size. Similarly, time for analysis scales adversely for power hungry Raman or surface abrasion, and also for low photon count gamma ray spectroscopy sensors. 1 cm scale rovers with MER-like science capability should be possible, and could operate on solar power out to 40 AU. At a 2 cm scale, rovers powered by current RTGs become feasible, and have the advantage of high data rates in the outer solar system.

Such vehicles will return images, autonomously navigate from interest point to interest point, and autonomously deploy analytical instruments. Analyses will include elemental compositions from APX/γ ray spectrometers and molecular and/or mineralogical information from Raman/infrared spectroscopy. This size scale is buildable with current engineering technology, as sizes of mechanical components are similar to watch movements. Low temperatures are not an obstacle. In contrast, very high temperatures (390-480°C for surface Venus) will require more work on high temperature electronics, but current developments suggest that dense, low power, integrated circuits for Venusian conditions should be feasible in the very near future. Smaller rovers of mm size, designed primarily for imaging are feasible at inner solar system solar flux levels, though if deployed in significant numbers, would probably require developments in manufacturing technologies to allow mechanised assembly of multiple MEMS systems. Miniature, degassing to space, RTGs could provide power in cold vacuum environments, as could advanced betavoltaic systems, though much work would be required to develop the latter.

The low mass systems described should allow preliminary surface exploration of multiple sites (10s to 100s )on most bodies in the solar system with a very low mass budget, perhaps as little as 10 kg per target planetary or planetoidal body.

## Abbreviations

**APX** Alpha Particle X-ray Spectrometer
**ASV** Adaptive Suspension Vehicle – vehicle scale heaxapod
**AU** Astronomical Unit ($1.5 \times 10^8$ km)
**BGO** Bismuth Germanium Oxide $Bi_4Ge_3O_{12}$
**BVV** Series of parallel processors using a small number of commodity microprocessors developed from the 70s to 80s
**CCD** Charge Coupled Device
**ChemFET** Chemical Field Effect Transistor
**CMOS** Complementary Metal Oxide Semiconductor
**CVD** Chemical Vapour Deposition
**Delta V** Delta Velocity (velocity change)
**Dof** Degrees of Freedom
**DSA** Deep space array
**DRAM** Dynamic Random Access Memory
**Eb/kT** Energy per bit/Boltzman's constant x absolute temperature
**III/V** Compound Semiconductor Based on a Period III and a Period V Element
**f** frequency
**FAST** Features from Accelerated Segment Test (corner feature detector)
**GCR** Galactic Cosmic Ray
**GTO** Geostationary Transfer Orbit
**IMPATT** (IMPact Avalanche Transit Time)
**IR** Infrared
**JFET** Junction Field Effect Transistor
**JPL** Jet Propulsion Laboratory
**LED** Light Emitting Diode

**LEO** Low Earth Orbit
**LIBS** Laser-Induced Breakdown Spectrometer/Spectroscopy
**LRV** Apollo lunar rover
**MEMS** Microelectromechanical Systems
**MER** Mars Exploration Rover (Spirit and Opportunity)
**MESFET** Metal Semiconductor Field Effect Transistor
**MIP** Million instructions per second (assumed to be 8-16 bit fixed point)
**MIT** Massachusetts Institute of Technology
**MSL** Mars Science Laboratory (Curiosity)
**OF** Optical Flow
**PV** Photovoltaic
**QCL** Quantum Cascade Laser
**RTG** Radioisotope Thermoelectric Generator
**SEU** Single Event Upset
**SOI** silicon on insulator
**SVM** Small Vision Module (SRI)
**UV** Ultraviolet
**XR** X-ray
**XRF** X-ray fluorescence
**Z** Atomic number

**Contents**

**Introduction**

Current space exploration is highly mass limited, due to launch costs of about $10,000 kg$^{-1}$ (e.g. [1,2]) placing severe constraints on probe size. However, space probes are information gathering systems and, as such, should be amenable to a high degree of miniaturisation. This would allow for a large number of missions to be launched, providing the ability to simultaneously explore multiple sites on a planetary body, and do this concurrently for multiple bodies in the solar system. The strategy has been suggested before by Brooks and Flynn (e.g. [3]). They noted that late 1980s robot technology allowed multiple, kilogram scale or smaller, robots to autonomously explore a planetary surface, with the primary target being either the Moon or Mars. This was due to a combination of the development of microprocessors and the subsumption architecture. The latter was composed of a hierarchically arranged set of simple reflexive behaviours which required minimal computational resources, and was fairly easy to expand. They envisaged hundreds or thousands of small robots exploring in parallel, with the entire swarm launched for the mass cost of a single Viking lander. They also raised the possibility of mm-scale, gnat sized robots [4,5]. JPL built a series of rover prototypes (Tooth and early Rockies) based on the subsumption architecture [6]. Since then, many authours have demonstrated small autonomously navigating robots using wheels, legs and tracks. Very small semi autonomous flying vehicles have also been developed, with weights in the 10s of g (e.g. [7]). Additionally, the subsumption-style rovers led to the Pathfinder microrover, and to the 1 kg scale Muses C rover [8,9].

This work will examine a range of constraints on rover miniaturisation, and their effects on mission design. The constraints examined will be level of autonomy, communication, power requirements, sensors, computation, thermal management, mobility and target specific issues, e.g. the high radiation at the surfaces of the inner Galilean satellites. High temperatures are examined in detail, as compositional data from Mercury and Venus will be vital in testing planetary evolution models. Present day, or very near term technologies addressing miniaturisation constraints are reviewed. Data transmission will be to an orbiting relay satellite which will send the data on to Earth (Fig.1a). The rover examined will be wheeled, communicate with an actively pointed dish antenna or equivalent, and will be equipped with a number of scientific sensors including some which are positioned with a manipulator (Fig. 1b). Each section has an introductory summary covering the major points.

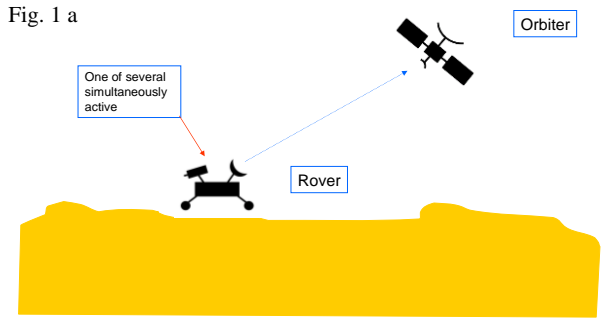

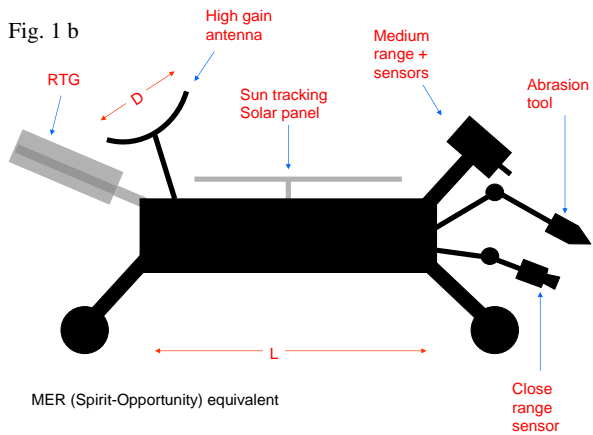

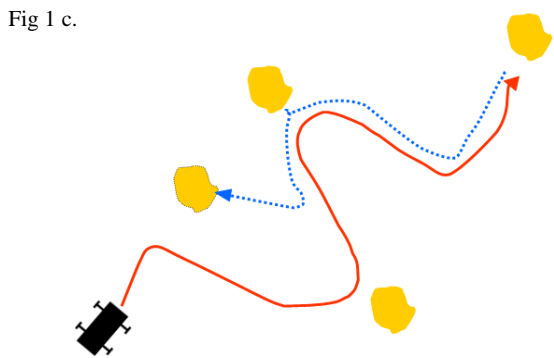

*Fig. 1a shows the basic mission architecture being considered. Fig. 1 b shows the basic rover components. Fig. 1 c shows the rover exploration strategy*

**Constraints**

*i) Autonomy and computational constraints*

*Uncontrolled exploration in a novel environment is wasteful of rover and communications resources. Remote control becomes difficult when multiple rovers are involved. Autonomous navigation and control of manipulators and instruments is feasible with, by modern standards, very limited computational resources (~10 Mips/ 10 Mbits, or an order of magnitude less with a subsumption approach). The best trade-off for a multi-rover scenario is semiautonomous operation, where navigation and manipulator pathplanning are automated, but science analysis and high level goal selection are terrestrially planned.*

So what are the constraints on exploration using miniature rovers? A primary consideration is the level of rover autonomy. The reactive subsumption architecture rovers of the late 1980s, were described as fast, cheap and "out of control", randomly exploring an environment. These required fewer than 256 logic gates (~1000 transistors), running at a few hundred Hz, to produce exploratory type behaviours [10]. On the other hand, the large MER rovers, and the MSL rover, have been too precious to entrust to much autonomous operation, and most traverses and analyses have been choreographed on Earth. This is feasible for a small number of rovers, but becomes more burdensome when considering tens to hundreds of rovers, and also for distant targets, where signal transit times will be considerably longer than for Earth-Mars-Earth communications. But what of the idea of a swarm of autonomous rovers fanning out across a planetary surface? The problem here is that the environment *will be entirely new to science*, resulting in rovers having to send large volumes of data back as they randomly explore, and targeting of Earth identified features of interest will be expensive in terms of rover numbers. The endmember of randomly exploring rovers will sacrifice the gains from small size and simplicity for large communication bandwidths, and inefficiency in targeting Earth identified interest points.

A compromise is to utilise semiautonomous navigation. This would involve the rover transmitting an image of its current environment, from which interest points could be identified on Earth. These could be sent back to the rover, which it would then autonomously travel to, and examine with its instruments. The examination of target rock/soil would be autonomous in terms of manipulator path planning and scientific instrument deployment. If conducting a more random style of exploration, it would transmit waypoint images and scientific data, and if these proved unusual, would be able to backtrack on command from Earth, and examine these areas in more detail (see Fig. 1c). Such a strategy would limit returned data to local area snapshots, interest point images, and instrument data. Commands would be at a high level e.g. navigate to point x,y in an image, and could be sent at low data rates. Many of the rover scientific instruments require tens of minutes to hours to provide an accurate composition. This fits well with an Earth command, autonomously navigate, analyse and transmit exploration paradigm. Such semiautonomous exploration has been within the robotic state of the art for some time (e.g. Rocky 7, [11]). With a large number of small inexpensive rovers, loss due to navigational errors or exploration of high risk terrain can be tolerated.

Examining the robotics literature, a rover moving at 0.1 rover lengths a second (0.1 L) would require about 10 Mips (table 1 gives the computational requirements of a variety of robots). This assumes low resolution 64x64 stereo (e.g. [12]), and visual feature tracking (e.g. the FAST corner detector [13], or the even simpler edge point feature of [14]) for odometery, navigation and targeting at ~1 Hz. This is similar to the Rattler stereo sensing rover prototype [15]. Active sensors using structured light or proximity sensors would lower costs to 0.1-1 Mips, as illustrated by the early Rocky rovers [6]. Real time, sensor controlled operation of a 6 d.o.f. manipulator requires ~1 Mip (e.g. [16]). The memory required for the control software would probably be ~ $10^7$ bits (see Table 1 for examples). Low power optimised, modern silicon technology, has 8 bit multiply energies of ~$10^{-14}$ J [17], SRAM memory cell leakage power loss of~ $10^{-13}$-$10^{-14}$ W/bit [19], and access energies of $10^{-13}$ J/bit (e.g. [20]). However there is a complex trade off between access and leakage energy in SRAM design [21]. Use of read only memory, where appropriate, can lead to significant power savings (e.g. [22]). Special purpose systems can improve efficiency by a factor of $10^2$ or more (e.g. [23]), due special purpose logic's increased efficiency, and reduced memory accesses for intermediate steps. Such specialised logic can be powered up or down as required. This suggests that a microrover's energy requirement for computation will be ~1 µW ($10^7$ bits in leakage optimised SRAM, a ~$10^4$ transistor, low power, 1 Mip processor and a small hardware accelerator for the computationally intensive parts of the navigation algorithms). Unless very extensive use is made of special purpose hardware, the silicon area will be dominated by memory. Current memory has a size of 10-100 $F^2$ (where F is the lithographic limit of the manufacturing system) per bit. At the present limit of ~20 nm, $10^7$ bits will occupy 0.2x0.2 mm to 0.6x0.6 mm. This could be reduced by stacking of thinned chips.

Subsumption style architectures may be able to give lower computational requirements (~ 1/40 processing and 1/50 memory) for a Rocky style rover (see Table 1). If built using special purpose logic, as few as 1600 to 5000 transistors might be needed, switching at perhaps 100 Hz (based on the MIT gate array controlled robots Tom and Jerry [10], and Tooth/Rocky III style rover memory usage). Extremely low power consumption would be feasible if implemented in a low leakage CMOS or other complementary logic – possibly at the subnanowatt level.

| | **Robot** | **Speed Mips** | **Memory Mbit** | **Comment** |
|---|---|---|---|---|
| **Sensor controlled arms** | Oracle[16] | 0.7 | 0.52 | |
| | Mitsubishi arm[23] | 1 | 0.09 | |
| **Legged vehicles** | Genghis[24] | 1 | 0.088 | Subsumption Arch. |
| | Titan III[25] | 2 | 0.5 | |
| | ASV[26,27,28] | 33 | 31 | Large vehicle |
| **Wheeled rover prototypes** | Tooth[6] | 0.5 | 0.028 | Subsumption Arch. |
| | Rocky 3[6] | 0.25 | 0.08 | Subsumption Arch. |
| | Rocky 4[6] | 0.25 | 0.16 | Subsumption Arch. |
| | Muses C[8] | 1 | 24 | |
| **Spacecraft control and sequencing** | Minuteman[29] | 0.013 | 0.2 | |
| | Saturn[30] | 0.01 | 0.4 | |
| | Skylab attitude control[31] | 0.06 | 0.26 | |
| | Viking lander[31] | 0.115 | 0.47 | Doppler radar |
| | Buran[32] | 0.4 | 26 | Ascent, orbit, descent |
| | Shuttle[33] | 1 | 9 | Ascent, orbit, descent |
| **Micro UAVs** | OF 30g Powered aircraft[34] | 5[35] | 0.27 | |
| | OF 10 g powered aircraft[7] | 8[35] | 0.6 | |
| **Vision systems** | BVV[36,37] | 1 | 1 | 60 Hz tracking* |
| | SVM[38] | 65 | 0.65 | 8Hz ~100x100 stereo |

*4 contour elements

*Table 1* gives an indication of computational requirements for a variety of robotic tasks. Sensor controlled, kinematically complex, 6 dof+ arms have been built using less than 1 Mip and 1 Mbit. Wheeled rover prototypes show similar trends and, when built using subsumption style architectures, utilise 0.1 Mbit or less. Legged vehicles capable of surmounting rugged terrain with a roughness of ~ 0.3 rover height can also be built under similar constraints, though the large, fast moving (in terms of leg movements), and kinematically complex ASV required an order of magnitude more computation. It also included a 100x100 laser ranging system for assessing traversabilty. Control of spacecraft ascent, attitude and landing requires surprisingly little computation, again with less than 1 Mip and ~0.5 Mbits. Even the much more sophisticated Buran and Space Shuttle required only 20 and 50x as much memory. Some idea of contour tracking requirements which could be used for features or rover track following are given by the BVV system which was used for high speed road feature tracking, while the small vision system 8 Hz 100x100 stereo depth maps needs ~50 Mips.

Despite low voltages and small device sizes, tolerance to radiation upset is high. Software error correction together with redundancy and voting will be suitable for much of the solar system [40], the major exception being the inner Galillean moons (see section vii).

Additional benefits can be gained from the ability to deploy multiple rovers. For example, precision local navigation can be simplified if two visually distinctive rovers are used, which can be easily segmented from the background terrain. If one rover remains stationary, the other can use it as a reference for distance and direction travelled. By alternating roles, precision navigation is possible using simple vision algorithms.

For the purposes of the majority of this study, a minimal computational electrical power requirement of 1 µW is assumed, which should be sufficient to support the 10 Mip/10 Mbit rover if some special purpose hardware is included to boost energy efficiency. Lower power requirements are possible for some simpler mobile rover designs.

*ii) Communications*

*Communication is limited by the Friis equation. Data rate is proportional to transmitter power and square of antenna dish diameter and frequency. Very high frequency operation can reduce power requirements, but is limited by falling electronic efficiencies and dimensional tolerances of the orbiter receiving antenna. Given current electronics, a reasonable compromise is 300 GHz. Science and navigation needs sets a minimum data rate of ~20 bits/s/rover. For a 1cm rover – 1m orbiter dish at $10^4$ km 2 µW are required.*

The primary task of a rover is to gather scientific information. This needs to be transmitted to Earth, and commands need to be transmitted to the rover. For the previously discussed rover strategy, a likely minimum field of view size is $6.4 \times 10^4$ pixels (256x256x8 bits, lossy compression to 1 bit/pixel - e.g. [41]) Addition of compositional information (point spectral analyses, ~100 channels at 10 bits per channel), suggests a minimum data volume of ~$7 \times 10^4$ bits.. Assuming one locale was targeted per hour, matching well with communication time lags and instrument analysis times, a data rate of approximately 20 bits/s per rover would be needed. Several microrovers are likely to be operating in parallel. For an assumed baseline of 5 simultaneously operating rovers it would rise to 100 bits per second.

The Friis equation (Eq. 1) indicates that the data rate is proportional to transmitter power and the square of receiver and transmitter aperture, and of frequency. Large transmit and receive dishes, and high transmitter power maximise data rate. A small vehicle could use a shorter wavelength to offset some of its dish size disadvantage.

Data rate $\alpha\ P_{Tr}\ D_T^2\ D_R^2\ f^2\ R^{-2}$ *(1)*

$P_{Tr}$ –Transmitter power, $D_T$ – Transmitter diameter, $D_R$ – Receiver diameter, f – frequency, R – Distance between transmitter and receiver

The communications system is assumed to be split between a terrestrial receiver and transmitter, and a relay station orbiting the target body. The distance of the relay orbiter to Earth is much greater than the distance of the orbiters to the landers, leading to an orbiter that is larger (in the case examined, 1 m scale) and smaller rovers.

Based on Earth-Mars X band (8 GHz) data transmission (100 W, 2.5 m transmitter to 34 m receiver, 2.6 AU, 300 kbits/s - [42]), and assuming the orbiter is $10^4$ km from the rovers and a shorter Ka band (30 GHz), a 1 cm transmitter requires 0.23 mW at 50% efficiency power to radiowave efficiency to transmit 100 bits/s. A link analysis based on bit energy/kT (Eb/kT) limit with a signal to noise ratio of 2 (easily achievable with modern error coding), a 300 K planetary background, and 75% antenna efficiency gives a 100 bits/s transmit power of 0.5 mW (see Table 2). The discrepancy between the two estimates is largely due to the difference in background temperatures (The DSA in [42] will have a sky view temperature of 50-60 K). It would seem straightforward to simply increase the frequency further, and benefit from an $f^2$ increase. Unfortunately things are not so simple. Raising the frequency to 90 GHz will reduce amplifier efficiency from ~50% to ~25 % (e.g. [43]), to 10% at 190 GHz and 5% at 270 GHz [44]. However, DARPA feels that 100 GHz plus amplifier efficiencies of 65% are achievable [45].

An IMPATT (*Imp*act *A*valanche *T*ransit *T*ime) diode source would have higher efficiency. Si-SiC IMPATT diodes can operate at about 25 % efficiency to 500 GHz [46]. With GaN, frequencies of 5 THz at 6% efficiency are achievable [47].

| | | |
|---|---|---|
| Wavelength | 1 | cm |
| Transmitter diameter ($D_T$) 75% efficient | 1 | cm |
| $Gain_{Tr}$ | 7.4 | |
| Receiver diameter ($D_R$) 75% efficient | 1 | m |
| $Gain_{Rec}$ | $7.4 \times 10^4$ | |
| Range | $10^4$ | km |
| Power Transmitter ($P_{tr}$) 2mW electrical | 1 | mW |
| Free space loss (FSL) | $1.6 \times 10^{20}$ | |
| Power received (Pr) | $3.46 \times 10^{-18}$ | W |
| Receiver antenna temperature ($T_{rec}$) | 300 | K |
| kT ($N_o$) | $4.2 \times 10^{-21}$ | |
| Maximum bit rate ($E_b/N_o$) | 825 | Bits/s |
| Bit rate at SNR of 2 | 413 | Bits/s |
| Bit rate at 1 mW electrical | ~200 | Bits/s |

*Table 2 shows a link budget for a 1 mW electrical, 50% electrical to radio efficiency, 1 cm antenna diameter, 1 cm wavelength link to a 1 m orbiter dish. The receiving antenna temperature is determined by the planetary surface temperature which the orbiter antenna is viewing, ranging from 750 K for Venus and Mercury surfaces to 40 K or less in the outer solar system. 300 K is used in the example above.*

Assuming ~270 GHz, with literature amplifier efficiencies and a 10 mm antenna, gives a power requirement of ~60 μW for 100 bits$^{-1}$. The orbiter-rover distance of $10^4$ km is only the case for a large body such as Venus or Mars. For a smaller body, such as a 1000 km icy moon, it could be reduced to 2000 km with a transmit power of ~2.5 μW. Using IMPATT efficiencies, 10 μW would be needed for 300 GHz and $10^4$ km, while as little as 150 nW would be required for a 5000 GHz source. However submillimeter wavelengths may be problematic for 1 m orbiter dishes, as thermal expansion/contraction may cause distortions that are a significant fraction of the wavelength, reducing gain.

For communications, a frequency maximum of 300 GHz will be assumed, with an electric to radiowave efficiency of 25% based on the recent work on IMPATT diodes. This gives a transmitter power requirement for a 1 cm rover dish to a 1 m orbiter dish of 10 μW at $10^4$ km and ~0.4 μW at 2000 km, based on the link analysis data rate at 300 K.

*iii) Locomotion.*

*At a speed of 0.1 rover length per second, solar power is adequate out to 30 AU, assuming either jumping\* rovers at a small scale/low gravitational field, or wheeled rovers at the larger end of the scale spectrum. Even at a size of 1 mm a rover could travel 3 km/yr, comparable to MER exploration rates. A 1-10 mm scale rover can easily navigate a planetary surface which is composed of sand sized material strewn with cobbles or boulders. Kg-scale clawed robots can climb moderately rough vertical and inverted surfaces at 1 G. Scaling of the energetics legged and wheeled rovers is complex, and requires more work to fully understand.*

*\*for clarity, a jumping rover jumps, lands, comes to rest, reorientates and jumps again. A hopping rover moves in bounds, does not decelerate appreciably on landing, and uses elastic energy recovery to maximise efficiency.*

For simplicity, only wheeled rovers will be considered in later sections. However other modes of locomotion have advantages. Small legged vehicles with appropriate gripping surfaces on their feet will be able to scale vertical surfaces. Microtextured polymer feet have allowed small terrestrial robots to scale smooth vertical surfaces [48]. These are susceptible to clogging with fine dirt, and may become unusable in a dusty planetary surface environment. A better option might be small claws which are able to grip minor asperites [49]. These have allowed 0.4-3 kg robots to ascend vertical rough concrete, stucco and tree trunk surfaces [49,50], and a metre scale, 7 kg, four legged crawler to move upside down on a rough basalt. [51, 52]. Gripping feet will also be useful in very small body exploration, due to the very low surface gravity, and will prevent the rover bouncing into space as a result of reaction forces from locomotion or instrument deployment.

Jumping robots are feasible at low masses (where the small size also gives better tolerance of impact due to square-cube effects). Using a polymer composite spring system (energy density~25 Jkg$^{-1}$) a vehicle has a jumping height of 2.5 m at 1 gravity. (Elastomers have higher energy densities, but are limited to a -100°C to 250°C). A 10 g, jumping robot using this design philosophy has been designed which can jump to a height of 76 cm at 1 gravity [53]. The height jumped (and distance traversed) will scale in proportion to the inverse of the local gravity field.

A metre scale wheeled vehicle at 1 G surface gravity, on firm sand or gravel surfaces, uses ~ 0.5 Jkg$^{-1}$m$^{-1}$. Scaling is not straightforward (Fig. 2). Jumping is constant in energy requirement at the scale considered (~3.5 Jkg$^{-1}$m$^{-1}$ at 1 G). Animal legged walking shows an L$^{-1}$ dependence, from about 1 g to several hundred kg, and then appears to become scale independent from 1 g to 1 mg (at 30 Jkg$^{-1}$m$^{-1}$). The scaling implies a complex combination of factors including gravity, elastic hysteresis losses and leg length. Similar scaling can probably be anticipated for legged mobile robots which have been optimised for energy recovery using elastic elements. Pneumatic tyred vehicles from bicycles to large road trucks have a relatively flat specific energy, but wheel diameter does not change dramatically with scaling. Some smaller vehicles have been plotted, and they appear to show similar scaling to legged animals, though it should be noted, that Rocky 7 was designed for rough terrain traversal, and Alice for ease of assembly. Neither was aggressively optimised for energy efficient locomotion. Similar complex interrelations to those seen in legged animals between gravity, hysteresis, soil penetration and design are likely.

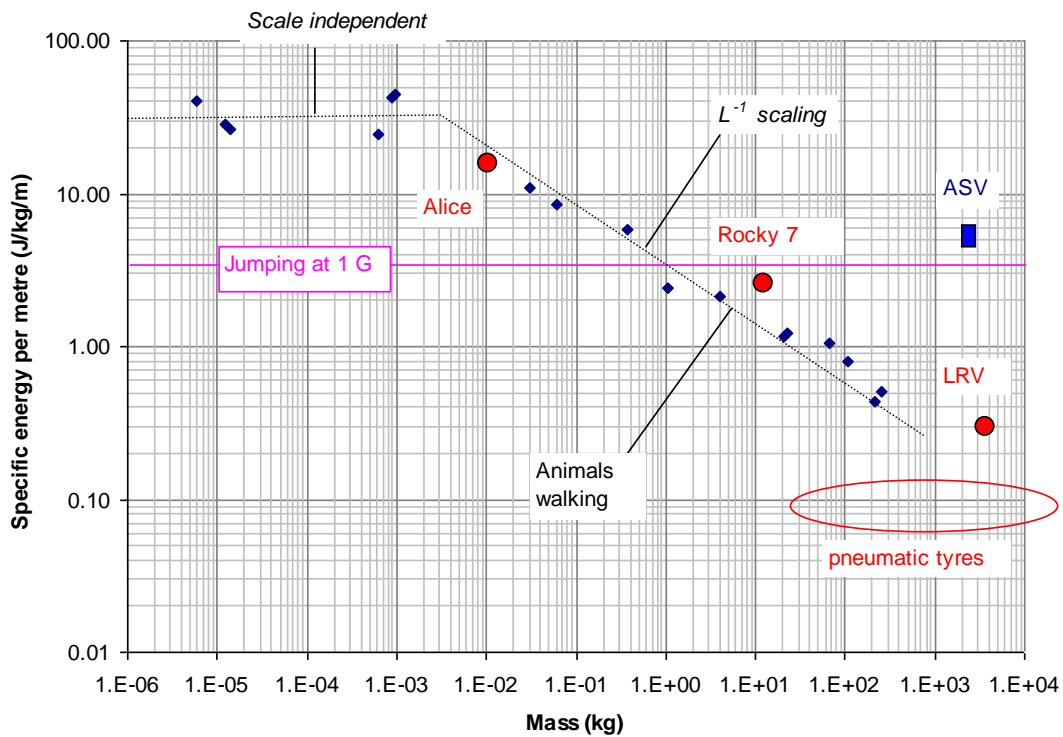

*Figure 2. The scaling of locomotion specific energy in Jkg$^{-1}$m$^{-1}$ for various animals and vehicles. The animal specific energies are from [54, 55, 56, 57]. All animal specific energies have been converted from metabolic energies derived from $O_2$ consumption/$CO_2$ production to muscular mechanical energy assuming a thermal efficiency of 25% (ATP production 50% efficient, ATP hydrolysis to muscular mechanical power 50% efficient). This approximates electrical power requirements for electrically driven vehicles. The robots are from [27, 58, 59].*

In view of the complexities discussed above, a locomotion energy of 1 Jkg$^{-1}$m$^{-1}$ has been selected to represent small jumping robots and larger wheeled rovers for energy limited designs. This energy is pessimistic for low gravity environments. Solar power is adequate for locomotion needs out to 40 AU. Even at a 1 mm scale size, a distance of 3 km could be explored in one year. Larger rovers allow greater exploration distances (300 km at 10 cm).

Actual maximum traverse distances will be limited by wheel bearing life, which is a complex function of abrasive dust ingress, bearing design, bearing seal life, and dust/bearing chemistry. Scaling predictions are therefore difficult. If dust ingress into bearings proves to be a major problem, a number of design options are worth investigating. Legged rovers with gaiters/boots are one option, though high performance elastomers are limited to -100°C to 250°C. Metal corrugated tubes could possibly be used at higher or lower temperatures – though current tube designs tackle different problems requiring less flexibility and high internal pressures. An alternative would be to use an internally actuated spherical design which can move and steer with no external moving parts. [60] reviews actuation mechanisms, and space applications include [61, 62, 63].

Typical planetary landscapes (Moon, Mars) tend to show a mixture of fine grained, <1 mm material with a greater or lesser density of scattered cobbles and boulders. A 1-10 mm scale vehicle would travel on the fine material, and manoeuvre round the cobles and boulders. The limited samples of Venusian terrain show evidence of exfoliation-like weathering with cm scale steps. This suggests a minimum Venusian rover length > 3 times the likely step height for an articulated multiwheeled rover. It should be remembered, that for Mars and the Moon, there is a strong selection bias, as missions have tended to target topographically low risk terrain.

*iv) Sensors and instruments*

*Imaging sensor size is diffraction limited, giving a sensor size of ~0.3 mm (λ 0.4-1 μm, 250x250 pixels). Small startrackers of ~ 200x200 μm scale appear feasible, and these, or low resolution directional antennas, would provide suitable angular information for antenna pointing. Elemental composition can be obtained using small APX (alpha particle X ray) or γ spectrometers. Absolute APX scaling is limited by the need to shield the excited XR sensor from the radioactive source – giving a minimum device size of 2-4 mm. In a vacuum environment modern silicon drift detectors are suitable for analysis of B onwards. Natural γ detectors are probably limited to a minimum size of ~2 $cm^3$ and 15 g due to the poor absorption of 1-3 MeV gamma rays, if a 24 hr analysis window is being considered. Development of new spectroscopic high Z detector materials such as PbO and $Tl_2S$ may reduce this to ~ 1 $cm^3$ and 10 g.*
*Mineralogical or molecular compositions can be obtained using Raman (>8 mm) or 1-20 μm infrared spectroscopy (>8-16 mm). Size is limited by the need to ensure adequate spectral resolution. Raman spectroscopy is also limited by laser power supply. Smaller infrared instruments could be created using a tunable IR source, or an array of narrow band IR sources. Quantum cascade lasers (QCL) offer a possibility, but current designs have a maximum length of 4 mm, and are power limited.*
*Microfluidic systems offer the possibility of wet chemical analysis at very small scales, but are limited by the useful operating range of silicone elastomers to -100-+250°C, solvent liquid ranges and sample size statistics. Mass spectrometers and LIBS type systems appear to be limited to sizes greater than 100 mm.*
*An abrasion system may be required to remove surface alteration. The size is power limited.*

A key question in assessing the viability of miniature rovers is the degree to which their sensor systems can be scaled and still provide a useful scientific return. With a minimum size of 250 x 250 pixels, the Rayleigh criterion (Eq. 2) gives a minimum lens diameter of 300 λ.

$(\sin)\theta = 1.22 \lambda/D$ *(2)*

$(\sin)\theta$ – Minimal separable angle in degrees (at small angles sin θ~θ), λ – Wavelength, D – Diameter of optics

With a short focal length giving a 90 degree field of view, the entire imaging package will occupy a minimum dimension of ~300 λ. The length of the imaging system may increase by ~2 or more if near field focussing optics are included. (Super resolution techniques can achieve finer resolution than eq. 2, at the expense of multiple images and high levels of computation – the latter limiting use on small rovers).

For the shortest wavelengths of UV to which common optical materials are reasonably transparent (LiF 105 nm+, $MgF_2$ 115 nm+, $CaF_2$ 125 nm+, [64]), the minimum imaging system size would be ~40 μm. However solar UV is, at this point, non black body and is derived from corona and flare processes. It has an intensity of ~ $10^{-4}$ to $10^{-5}$ of visible/near UV [65], but with a strong H Lyman α emission peak at 121 nm [66]. High thermal solar UV levels begin at ~250 nm. Size limits for visible light the size would be ~200 μm, for near infrared (to ~3 μm) 900 μm, and for thermal infrared ~ 3-6 mm. Imaging array powers are low and arrays have been designed to operate at the μW level (e.g. [67]).

As energy efficient communications requires directional beams, rover position relative to the orbiter is needed. This can be done using, sun, star and, where present, planet/moon trackers. Star tracking represents the limiting case, as photon fluxes from other sources are orders of magnitude higher. Star trackers are constrained by target photon flux, sufficient detector array density/optical resolution to separate target stars, and a focussed target photon count rate greater than the night sky background of $10^2$ photons $μm^{-2}s^{-1}$. Photon flux from the 15 brightest stars at ~ 0.55 μm is ~0.03 photons $μm^{-2}s^{-1}$. At a detector element size of 1x1 μm and a pixel number of 200x200 giving a system size of 200x200 μm, the tracker would detect ~1,000 photons $s^{-1}$ per target star (vs. a background count per target star of 100 photons $s^{-1}$), which is enough for a position fix for a stationary rover. An alternative strategy would be a low resolution sensor (size ~ λ) to detect and track the orbiter's communication beam. On a short time scale rover attitude can be determined using MEMS accelerometers and angular rate sensors. Angular rate sensors are now sub mm in size (e.g. [68]). Analogue to digital converter sizes can be as little as 35x35 μm for a 20 nm process (e.g. [69]), with energies of ~10 fJ per sample [70] and so do not represent major scaling limits.

A primary objective is obtaining compositional information about rocks, soils and fluids near the rover. A variety of small scale instruments are available to derive elemental or molecular/mineralogical composition. Historically elemental composition from Na onwards has been determined using alpha particle XR spectroscopy (APX) [71, 72, 73]. α and XR excitation of the specimen surface excites XR emission/fluorescence This is then detected with an energy dispersive semiconductor detector. Lower Z element analysis was limited to α backscatter analysis. Modern detectors have improved, and silicon drift detectors can spectroscopically detect X-rays emitted by B onwards at operating temperatures of -20$^o$C [74], which is above ambient temperatures in many target environments.

40 mm scale APXs have been flown on the MER rovers [72]. Analysis are on the order of 5-15 of minutes for parts per 1000 analyses [73], but ~ 10 hrs for backscatter. A 20x20 mm scale, XRF only, version has been flown on Curiosity, and utilises the same XR sensor and α sources as the MER devices, but is only 20 mm from the surface [73]. It is three times more sensitive. Scaling of count rate appears proportional to $L^{-2}$. Closer examination of the design reveals that the XR sensor intercepts only 0.2% of the available XR flux. A more geometrically efficient design with a 1 $mm^2$ source (~1/500 of the MSL source area), and a detector which is several times the source diameter, should intercept most of the available XR flux if held close to the target surface. This should offer an analysis time similar to the MSL APX. The principle limit on scaling appears to be the need to shield the detector from source X rays. This will probably limit device size to the 2-4 mm region.

A limiting factor for APX is the presence of a dense atmosphere. This is most problematic on Venus (90x Earth density), requiring unfeasibly close contact distances for α excitation. Reliance on XR fluorescence will be needed, which has a low efficiency for Z < Fe. However α excitation of atmospheric gases will produce a useful signal for higher Z element containing volatiles such as Ar, Kr, Xe, $SO_x$, HCl, $PH_3$, $AsH_3$, and $GeH_4$. $SO_x$ $H_2SO_4$, and HCl are likely in the Venusian atmosphere, while the high Z element hydrides and heavy noble gases are targets in gas giant atmospheres.

Another method of elemental analysis which is feasible for small rovers is natural γ/GCR excited γ ray spectroscopy. It allows measurement of surface abundance naturally occurring radioactives ($^{40}$K, U and Th). In addition, where a thick atmosphere is absent, GCR excitation of nuclei is significant, and can be used to derive compositions. Detectable elements include, H, C, N, O, Na, Mg, Al, Si, S, Cl, K, Ca, Ti, Cr, Fe, Th and U [75] (see table 3). Detection is energy dispersive, using a high Z semiconductor or scintillator. Surprisingly small semiconductor detectors are possible. These were developed when the properties of the relevant semiconductor crystals were poor and only mm scale detectors were feasible. For less than 1 MeV GCR induced γ rays, active volumes of 3x3x1 mm may be sufficient [76]. Efficiencies for such small devices were in the 1 to 10 percent range. A 2 $cm^3$ CdZnTe sensor has an efficiency of 7.1% for 662 KeV gamma rays [77]. Certain high Z detector materials have poor charge carrier collection efficiency if the crystal size is large. This can be overcome, with minimal energy resolution loss, by using stacked detectors (e.g. [78]). Unfortunately 1-3 MeV gamma rays are highly penetrating (see Fig. 3), even in high Z materials (~ 30 gcm$^{-2}$ half distance, excluding high angle Compton scattering). Several important lines (e.g. $^{40}$K, see Tables 3 and 4) occur at this energy. However, this does mean that γ spectroscopy can sample to a depth of several tens of cm in typical regolith materials.

γ ray absorption physics is complex. Primary absorption occurs via the photoelectric effect at lower energy and high Z, by Compton scattering at intermediate energy, and by $e^-/e^+$ pair production at high Z and energy. For thin spectroscopic detectors, photoelectric and pair production absorption are the main mechanisms to consider (the most probable forms of Compton scattering dissipate only a small amount of energy per scattering event, and the γ photon will either be absorbed by a photoelectric or pair production mechanism, or escape the detector). At low energies, photoelectric absorption produces a photoelectron with energy approximately equal to the absorbed γ photon. For a high Z detector this is important up to ~ 4 MeV. At high Z, pair production becomes important at γ energies > 1.1 MeV and, at 2 MeV, exceeds declining photoelectric absorption. The energy absorbed is converted into the rest mass of an $e^-$ and $e^+$, and the remaining energy is split between the $e^-/e^+$ kinetic energy. The $e^+$ will annihilate with an $e^-$ producing two 0.5 MeV gammas with a high Z photoelectric absorption half distance of 3.5 g/cm$^2$. The escape of pair production γ rays produces one or more escape peaks.

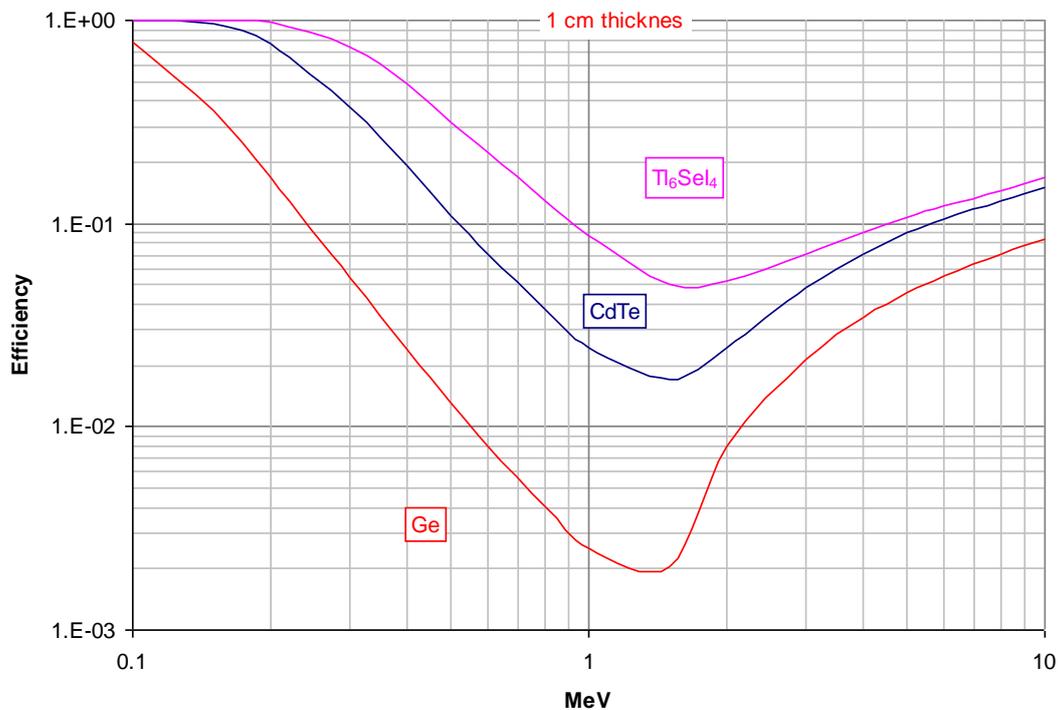

*Fig.3. illustrates the importance of high Z materials for approximate spectroscopic detection high energy gamma rays. Photoelectric and pair production absorption is shown for a 1 cm detector with normally incident γ rays. The effects of escape of less than 1-2 MeV photoelectrons is ignored, as is the escape of high energy, pair production $e^-/e^+$. Escape of 0.5 MeV $e^+$ annihilation gammas will produce distinct "escape peaks" for >1 MeV γ rays, which will still allow elemental identification.*

*Ge (5.3 gcm$^{-3}$) is a popular lower Z semiconducting spectroscopic material due to the high spectroscopic resolution of large, cryogenically cooled, ultrapure crystals. At a 1 cm thickness, it has 0.05 to 0.1 of the efficiency of high Z CdTe and Tl$_6$SeI$_4$.*

*Compton scattering has been ignored, but multiple scatters reducing energy to the point where photoelectric absorption is probable will tend to increase efficiency. This becomes an important mechanism for thick detectors of all Z values.*

In the case of a scintillation detector, γ rays produce UV-visible photons which have historically been detected with image intensifiers. However, PIN photodiodes and other solid state photodetectors can be used (e.g. [79]). They are more compact, and more appropriate for the small scale designs being discussed. Scintillators have lower energy resolutions than semiconductor detectors.

| Element | strongest γ MeV | Concentration ppm | Flux cm$^{-2}$day$^{-1}$ | % error (24 hr) 100% efficient | 10% efficient |
|---|---|---|---|---|---|
| K ($^{40}$K) | 1.46 | 1200 | 3456 | 2 | 5 |
| U | 0.61 | 0.5 | 1584 | 3 | 3 |
| Th | 2.61* | 1.9 | 3168 | 2 | 2 |
| Na | 0.44 | 3500 | 86.4 | 11 | 11 |
| Lu | 0.31 | 0.5 | 11 | 29 | 29 |
| Sm | 0.33 | 7 | 20 | 22 | 22 |
| Gd | 1.19* | 8 | 20 | 22 | 22 |
| Ni | 9 | 400 | 10 | 31 | ND$^+$ |
| Fe | 0.85 | 9x10$^4$ | 1656 | 2 | 8 |
| Al | 2.21 | 1.1x10$^5$ | 972 | 3 | 10 |
| Ca | 3.74 | 1.0 x10$^5$ | 504 | 4 | 14 |
| O | 6.13 | 4.35 x10$^5$ | 3744 | 2 | 5 |
| Si | 1.79 | 2.0 x10$^5$ | 4636.8 | 1 | 5 |
| Ti | 6.76 | 1.4 x10$^4$ | 593.28 | 4 | 13 |

*other abundant gammas are low energy, $^+$not detectable.

*Table 3. The effect of detector efficiency on spectra from a lunar regolith composition (24 hr observation, 1 cm$^2$/1cm$^3$ detector, 100%/10% efficiency). Major rock forming elements are observable at ~ 10% error with a small, high Z detector, which has a spectroscopic efficiency of 10% in the 1 to 2 MeV band. For lower energy gamma rays emitted by U, Th, Na, Gd and Lu detector efficiency remains high, and these can also be detected. Ni becomes undetectable with a 10% efficiency detector.*

Using typical lunar regolith [75], and assuming a 1 cm$^2$ detector and a 24 hr analysis period O, Na, Mg, Al, Si, K, Ca, Ti, Fe, Ni, Gd. Lu, U, Th could be detected assuming 100% detector efficiency. All except Ni could be measured at ~10% detector efficiency for high energy γ rays (see Table 3). For a small detector, spectroscopic count rate scales in approximate proportion to detector high Z element mass. As a large number of peaks to be discriminated, a high energy resolution semiconducting detector is probably best. For detection of 1-2 MeV gammas at ~10% efficiency, HgI$_2$ (ρ 6.4 gcm$^{-3}$, mass$_{10\%}$ 17g, band gap 2.1 eV) has the best high energy cross section of the more mature semiconductors, though high performance has been demonstrated for Tl$_6$SeI$_4$ (ρ 7.4 gcm$^{-3}$, mass$_{10\%}$ 13 g, band gap 1.9 eV) [80]. The best performance comes from a semiconductor with a high Z fraction, which paradoxically means that, for Pt to Bi, the more electropositive element should be low Z e.g. O or S. The best potential performance would come from Tl, Bi or Pb oxides or sulphides, requiring about 10-11 g and ~1 cm$^3$ of material for 10% detection efficiency. Many of these materials have been investigated as detectors for IR/radiography, with PbO (ρ 9.7 gcm$^{-3}$, mass$_{10\%}$ ~10 g, band gap 1.9 eV) having the best figure of merit. C, Si, Mg, N, S and O all have detectable γ energies in the range of 1-4 MeV, where absorption is at a minimum. A minimum thickness of 3.6 gcm$^{-2}$ ~ 6 mm of HgI$_2$, (see Table 4) is needed to capture γ rays from these major elements. This will also set the size limit in a high natural γ flux environment such as inner gas giant moons, and dayside Mercury in the presence of an active sun. Data in [75] does not include Li, B, Br or I which may be enriched in a Martian environment, and which are likely to have strong neutron capture related γ emission. High concentrations of these elements are seen in evaporites in various terrestrial arid and hyperarid environments [81]. In an icy moon or carbonaceous chondrite-like environment, N will be detectable, and probably C, as less probable emission modes could overcome interferences from O. Erupted and subsequently frozen brines would be likely to have a 'reduced' seawater composition, with the major elements being alkali metals and earths, halogens, with high Fe(II)), and also Co, Ni and Mn. Venusian environment problems are discussed in section vi.

Review of the mass spectroscopy literature shows a minimum system mass of ~1.5 kg (e.g. [82]). In a vacuum, mass spectrometers can be smaller as high vacuum pumping equipment is not required, and scaling is limited by the ability to fabricate a detector array and beam focussing. Acceleration energy will need to be large relative to the ambient ion thermal energy. In a vacuum environment a surface component will be detectable, due to solar wind sputtering and, in the inner solar system, thermal evaporation of more volatile substances. The degree of scaling achievable remains uncertain at present.

| γ energy MeV | | Photoelectric min t gcm$^{-2}$ | cm HgI$_2$ | Pair min t gcm$^{-2}$ | cm HgI$_2$ |
|---|---|---|---|---|---|
| 0.2 | **U** *F Mn Gd* | 0.2 | 0.03 | | |
| 0.3 | **Th** | 0.4 | 0.06 | | |
| 0.4 | *Ti* | 0.5 | 0.08 | | |
| 0.5 | **Na** | 0.7 | 0.11 | | |
| 0.7 | **Fe** | 1 | 0.16 | | |
| 0.8 | **Ca** | 1.3 | 0.20 | | |
| 0.9 | *Ni* **Al** | 1.4 | 0.22 | | |
| 1.1 | **Si** | 2 | 0.31 | 2 | 0.31 |
| 1.3 | (Cl) *P* | 2.6 | 0.41 | 2.6 | 0.41 |
| 1.4 | **Mg** | 2.8 | 0.44 | 2.8 | 0.44 |
| 1.5 | **K** *Ba* | 3 | 0.47 | 3 | 0.47 |
| 1.7 | [N] | 3.4 | 0.53 | 0.5 | 0.08 |
| 1.8 | **O** | 3.6 | 0.56 | 0.6 | 0.09 |
| 1.9 | *Sr* | 3.8 | 0.59 | 0.7 | 0.11 |
| 2.3 | (S) H | 4.6 | 0.72 | 0.9 | 0.14 |
| 4.5 | [C] | 7 | 1.09 | 3.5 | 0.55 |

*Table 4 shows the minimum energy of γ rays available for detection of a particular element, and the minimum detector thickness required. Significant γ sources in silicate rock are shown in bold, trace elements in italics. The latter will only be detectable with very long integration times and/or large detectors or unusual geology. S and Cl are strongly enriched in the martian regolith (curved brackets). H will be found in ices and hydrous minerals. N and C (square brackets) will be enriched in icy moon environments and, also for C, in carbonate environments. Major rock forming elements require an energy minimum of 1.5-1.8 MeV, depending on whether O or K is the cutoff, and a minimum detector thickness of 3-3.6 gcm$^{-2}$ – as photoelectron processes are dominant at this energy. This detector thickness would be suitable for up to 7 to 8 MeV, as at higher energy, pair production dominates for high Z materials, and would allow detection of S, H and C. Pair production at less than 1.4 MeV is very inefficient. At ~0.5 and 1 MeV interference occurs from $e^+$- $e^-$ annihilation gammas. Above this energy, thin detectors utilising $e^-/e^+$ kinetic energy are feasible, though efficiencies will be low, as the initial pair production cross section is still relatively low.*

Laser ionisation and subsequent optical emission spectroscopy (LIBS) has been employed on MSL. In a vacuum environment, time of flight spectroscopy is also possible [83]. Scaling is complex. Spectroscopy of emitted, 0.2-0.8 μm, light needs 3000 λ resolution [84]. Pulse time and peak power are a complex function of material volume and vaporisation energy, optics peak power tolerance, and the lasing material system. The smallest system found in the literature has a laser mass 60 g, L max of 100 mm and a systems mass of 200 g [85]. Due to the complexities of LIBS scaling a minimum size of 100 mm will be assumed.

Molecular or mineralogical composition can be determined by optical or infrared spectroscopic techniques, which are sensitive to bond strength and relative mass of bonding atoms. Two techniques are promising, near to thermal infrared (~1-20 μm) diffractive spectroscopy, and Raman spectroscopy. Infrared at shorter wavelengths can be used for pattern fitting to known minerals, while longer wavelengths allows detection of molecular stretching frequencies and, at slightly longer wavelengths, lower energy bending frequencies. Illumination for <3 μm infrared is solar dominated. In the inner solar system, thermal emission is the main source of >5 μm infrared. In the outer solar system ambient temperatures fall to levels where thermal, <20 μm, emission becomes negligible. Solar 10-20 μm radiation is ~1/1000 of visible and near infrared, compounding the low total intensity of sunlight. Active illumination will probably be required. Due to reliability issues with precision optical moving parts, Fourier transform spectrometers will not be considered, despite their high signal to noise ratio, and widespread terrestrial use. Present diffractive spectrometers are limited by grating size, optics diameter and size of detector element. Ideally a minimum of 100 channels is likely to be needed for spectroscopic infrared. This will give a minimum size of ~8x100 λ. Due to the increased optical complexity of the elements in a diffractive spectrometer, size in one direction is multiplied by 8. Minimum size for a near infrared spectrometer is ~2.5 mm, and, for thermal infrared ~ 8-16 mm. Examination of literature miniature spectrometers suggests that this is a reasonable limit [86]. One possible strategy to reduce the minimum size would be to utilise an active tunable infrared diode or laser light source, or an array of laser diodes spanning the spectrum of interest. The spectrum would be obtained by scanning the active source/array through its frequency range and measuring the infrared in each band. This could allow scaling well below the mm size. Quantum cascade lasers (QCLs) emit over narrow bands in the spectroscopic infrared. Such lasers are 5–20 μm wide and 1–3 mm long. Pulsed operation in the 5-12 μm region is possible at 300 K. Longer wavelength operation requires 100 K temperatures [87]. A 4 mW electrical input power integrated QCL has recently been described [88]. Continuous wave operation with imput powers of as low as 29 mW has been demonstrated [89]. Assuming an uncooled bolometer style thermal detector with a noise equivalent temperature detection of ~0.1 K and a maximum response frequency of 10 Hz, combined with 100 pulsed QCLs, and a 4mW input power for pulsed continuous operation, ~10 μW are required for a 100 band analysis over 1 hour, with a pulsed electrical energy storage requirement of 0.4 mJ. A 1 mg capacitor (at 0.36 $Jg^{-1}$ – see section v) would be needed to store power for 0.1 s of operation. This pulsed power storage would not be necessary for power systems generating 4 mW or more. Power requirements could probably be reduced if a detector with a faster response was used, in combination with shorter pulses. The minimum size of the system would be set by the length of the lasers, which is conservatively estimated at 3 mm (the smallest size found in the literature), and minimum power at 10 μW for a 1 hr analysis time.

In Raman spectroscopy the point being analysed is illuminated by a very intense monochromatic laser. The backscattered light contains a small (~ 1 in $10^5$), frequency shifted component. This reflects the type of bonds in a similar way to infrared spectroscopy. Energy per analysis is on the order of 0.1 J. Despite the high light intensity, minimal surface change occurs (Raman has been used to analyse historic works of art). As the total shift range is ~10% of the wavelength, the minimum size of a diffractive spectrometer will be ten times larger than a simple spectrometer operating at a similar wavelength. Due to fluorescence problems at short wavelengths, the minimum wavelength is likely to be 1 μm, giving an 8 mm minimum device size. At 10% laser efficiency, 0.3 mW are needed for a one hour analysis. Diode lasers need minimum operating powers of 1-10 mW to lase efficiently, even if 10 μm scale vertical lasers [90]. This high power is easily provided by capacitors, where 2.5 $mm^3$ of capacitor will provide 1 mJ electrical at a power 10 mW for 0.1 s or less (see section v). Raman and QCLs are assumed to be energy, as well as diffraction, limited.

Surfaces at most exploratory targets are affected by superficial alteration, due to either atmospheric weathering or solar wind implantation. Removal may be advantageous. Weathering/radiation damage will reduce the mechanical strength of this layer. Assuming 0.1 $Jmm^{-3}$ abrasion energy (this is a rather high estimate), 1.5 mW are needed to excavate a 5 x 5 x 2 mm deep hole in one hour. Some form of gripping of the target rock may be needed for small rovers and/or low gravitational fields to provide adequate reaction forces for grinding. Mars experience indicates some rock specific elemental variation between rover abraded and non abraded rocks [91]. This is due in part to operation in a more deeply penetrating XRF like mode for higher Z elements, and the depth of $Cl^-/SO_4^{2-}$ weathering crusts in old Martian surfaces.

One additional analytical strategy that needs to be touched on is 'wet' chemistry using microfluidics. Highly sophisticated microfluidic reactor and valve arrays have been developed, containing $10^3$ reactor chambers and $4x10^3$ microvalves in a 4 $cm^2$ integrated unit [92]. Microvalves are usually elastomer based, with pressure in a control line squeezing shut a deformable tube or membrane. Further scaling could be imagined, with channel widths of 10 μm, based on biological fluid channel sizes. Colormetric/conductivity/ChemFET systems could be used for detection/quantification. Molecular analysis could be assisted by integrated chromatography. The system would be limited by a supply of sample solvent and, depending on the analytic techniques used, other reagents. However, the small scale of the system would result in low reactant consumption, allowing at least tens to hundreds of analyses to be run. An important limitation is the nature of the solvent used for specimen digestion and transport. HF will digest silicates, but may cause problems with elastomer compatibility. Water, or other polar solvents, will dissolve typical evaporite minerals, such as those found on Mars. Non polar solvents can be used for non polar, low mass organics. Useful operating temperature will be limited by the freezing point of the solvent (or a eutectic containing the solvent). Systems using HF will need either a neutralisation step, or fluorinated non-silicone elastomers. The latter may have a reduced working temperature range. Design becomes more complex outside the temperature range of silicone elastomers (-100-250°C), as these are the key materials for flexible microvalves. Bubble based fluidics, as described in [93], could be used to overcome this problem. Thermal bubble generation is used, and high flexibility materials are not needed. However they are less well developed. Fluidic logic based on high Reynolds number effects, as was popular for control applications in the 60s and early 70s, does not scale to microsizes (small channels have very low Reynolds numbers - e.g. [94]).

Very small reaction chambers can be envisaged on the order of 100x100x10 µm. However the main size limiting step may well be digestion. 1 mm$^3$ or more may be needed for a representative sample, and to avoid the mechanical difficulties of handling very small amounts of granular material.

Sensor/instrument scale limits are summarised in table 5

| Sensor | Size | Notes |
| --- | --- | --- |
| **UV** | 30 µm | |
| **Optical imaging** | 0.175 mm | |
| **<3 µm IR imaging** | 0.75 mm | |
| **<3 µm IR spectroscopy\*** | 2.4 mm | |
| **Nat. γ min (major silicate elements)** | 5 mm (1 g) | |
| **APX** | 2-4 mm | Source shielding |
| **QCL** | 3 mm | 4 mW/10 µW |
| **Raman\*** | 8 mm | Laser power. ~0.1 J$_{optical}$/spectrum |
| **10-20 µm IR spectroscopy\*** | 8-16 mm | |
| **Nat. γ efficient** | 12 mm + (15g+) | |
| **Abrasion** | | 60 µW (24 hr excavation) |
| **LIBS\*** | 100mm+ | High pulsed power requirement |
| **Mass spectrometry** | 100mm+ | |
| **Microfluidics** | | **Liquid range** (HF eutectics for silicates, NH$_3$ eutectics for water ice) Sample size statistics |

\* *These systems are cylindrical with the maximum device dimension approximately 8 times the minimum.*

Table 5. Size limits of various sensor/analytical device technologies.

*v) Power*

*At <100 mm solar photovoltaic (PV), radioisotope thermoelectric generator (RTG), electrochemical battery/fuel cell and alpha/betavoltaics are candidate power sources. At 1 cm size, solar is viable for rover power to 40 AU (Pluto). Current RTGs have a minimum size of 2 cm and show a severe decline in efficiency as they approach this limit. Despite this, they are suitable for rover operation at this size. If vacuum insulated, and with a low emissivity coating, much smaller, relatively efficient, RTGs could probably be built down to a mm scale providing no atmosphere is present. Beta/alphavoltaic systems offer high theoretical power densities (similar to larger RTGs,) but this has yet to be reached in existing devices. Current systems compare unfavourably to RTGs under most conditions. Batteries and fuel cells offer viable long term power sources, providing temperatures allow ion mobilities (T>-50°C for current chemistries). Both are capable of extreme miniaturisation. High capacity thin film capacitors offer a useful secondary energy store which is capable of low temperature operation. At very low temperatures, high dielectric constant materials are available which boost energy density. Simple non operation during periods of non illumination is a viable option for cold night environments. Data rates scale as L$^5$ for radioisotope/electrochemical systems and L$^4$ for solar powered systems. For radioisotope power, Raman, abrasion and QCL sensors are the first to be limited by small rover size (6-10 mm). Power for communications becomes an issue at 4 mm.*

*Rotary actuators based on electrostatics, piezoelectrics, electromagnetics or magnetostrictives are suitable for use at cm-mm scale. Miniature piezoelectric actuators can show bending strains of up to 150$^o$, which would be useful for a variety of systems.*

Some indication of rover power requirements has been obtained from mobility, computational and communications estimates in the previous sections. Two primary long term power sources have been used; PV and RTGs (e.g. [95]). Some shorter lived missions such as Huygens and the Galileo Jupiter entry probe have used batteries [96].

Solar power relies on a line of sight to the sun – though it is possible to use reflected light from adjacent illuminated slopes. At 1 AU, multilayer GaInP, GaAs, and Ge panels can generate ~400 Wm$^{-2}$ at ~30% efficiency [97]. Specific power can reach 1200 Wkg$^{-1}$, at 12% efficiency with advanced thin film technologies [98]. Power falls with the square of distance to the sun, resulting in low mass efficiencies in the outer solar system. This becomes significant for solar powered rovers when array power drops to <0.5 µW, impeding communications to an orbiter, even at a 2000 km (see figs. 4 and 5). Low light intensities are not a major efficiency issue out to at least 22 AU (28% 1 AU, 25% 22 AU - [99]). Solar power will not operate in the dark so secondary energy storage issues become important (see later). Dust may be a problem, but can be dealt with by mechanical cleaning devices. In the case of the Mars rovers, while power loss has been significant, ambient Mars winds have removed excessive dust.

The alternative to solar power is radioisotope based thermoelectric or beta/alpahvoltaic power. Other radioisotope thermal-electric systems include thermophotovoltaics, sodium-β alumina thermal converters, thermionics and mechanical heat engines [100]. The first three require high hot end temperatures which are problematic at small sizes. Mechanical engines such as Stirling engines, while efficient, have the problem of complex moving parts. These systems will therefore not be considered further. In RTGs thermal energy from radioactive decay is used to drive a semiconductor thermoelectric generator. Efficiency is significantly less than ideal Carnot values, and less than heat engines with moving parts. However the solid state nature of the system ensures high reliability, and recent materials developments suggest that thermal efficiency, and hence power densities, may be raised by a significant factor.

The radioactive heat source of choice is $^{238}$Pu (0.5 W$_{thermal}$ g$^{-1}$, t$_{1/2}$ 88 yrs), as this has a very low emission rate of penetrating radiation, and hence low shielding mass. An important factor in the weight of radioisotope systems is the need to provide launch fire/high velocity reentry hardening. This consists of a refractory case. Typical power densities are 5 Wkg$^{-1}$, though the small RTGs formerly in pacemakers have lower power densities. The smallest $^{238}$Pu thermoelectric source found in the literature had a volume of 4.3 cm$^3$, a radioisotope source size of 5 mm long x 4.2 mm diameter (1.4 mW electrical, 150 mW thermal, ΔT 24 K). It is unclear whether the design was reentry hardened [101]. A recent design using microfabricated thermoelectric modules and a $^{238}$Pu heat source has a mass of ~ 60 g, a scale size of 4 cm, and a power output of ~50 mW [102]. Current small RTG designs are viable down to rover sizes of 2-3 cm.

An alternative method of energy conversion is via α or betavoltaics. Here instead of photons generating hole-electron pairs in a PV semiconductor system, α or β particles from a radioisotope are used. Alphavoltaic systems have problems with semiconductor damage, though this can be avoided by overlaying the source with a phosphor, and then using conventional PV energy conversion. Tritium ($^3$H, t$_{1/2}$ 12.3 years, 5.7 keV), $^{63}$Ni (t$_{1/2}$ 100 years, 17 keV), and $^{147}$Pm (t$_{1/2}$ 2.4 years, 73 keV) are the preferred β$^-$ isotopes due to their emission of pure, low energy, betas which cause relatively little lattice damage. For $^3$H betavoltaics typical power densities of ~ 0.55 µW$_{electrical}$ cm$^{-2}$ [103], and with 10 µm active layers, a power density of ~275 µWcm$^{-3}$ and 275 µWg$^{-1}$ may be attainable. [104] felt that 1000 plates cm$^{-3}$ might be overoptimistic and suggested 100 plates with a power density of 30 µWcm$^{-3}$

To estimate achievable, longterm, betavoltaic power density the following design was investigated. A GaN photovoltaic, with an assumed active layer thickness of 7 µm (two 3 µm GaN layers sandwiching 1 µm of MgT$_2$-the maximum thickness before self absorption of the 5.7 keV β$^-$ becomes a problem). A power density of 5 mWcm$^{-3}$ and 1 mWg$^{-1}$ is theoretically achievable. This compares with a $^{63}$Ni-GaN system proposed by [105] with an estimated theoretical power density of 6 mWg$^{-1}$. For long duration flights to the outer planets, the 12.3 year half life of $^3$H may present difficulties.

An optimised alphavoltaic systems might achieve a higher energy density [102]. A $^{238}$Pu based system utilising a GaN betavoltaic with a 5 μm PuO$_2$ layer sandwiched between two 20 μm GaN alphavoltaics would give a theoretical electrical power of 150 mWcm$^{-3}$ and 25 mWg$^{-1}$. This is without reentry hardening.

Extreme scaling of $^{238}$Pu systems will be limited both by the problems of maintaining high thermal gradients, and the need to shield from Pu K$_\alpha$ radiation which requires ~2 mm of high Z metal e.g. W (for x10$^6$ reduction in K$_\alpha$ dose). Alphavoltaic systems based on $^{244}$Cm or $^{238}$Pu will face similar issues. $^3$H based systems can be scaled to much smaller sizes due to the low shielding requirements of the 5.7 keV β$^-$ (range <1 μm in high Z materials). At a mm scale, a vacuum insulated $^{238}$Pu RTG with a low emissivity (0.05), polished surface should be viable, providing that the external environment was a vacuum. The radiative equilibrium temperature is ~ 300$^o$C. The beneficial effects of surface area to volume ratio on heat loss would allow a lower operating temperature at the heat rejecting radiators, and hence maintain a large temperature drop despite the low source temperature. A vacuum environment is necessary to allow venting of accumulated radiogenic $^4$He, and hence maintenance of vacuum insulation. Similar issues would arise for a $^3$H based RTG venting $^3$He. Power levels scaling with rover size, together with other constraints, is shown in Figs. 4 and 5.

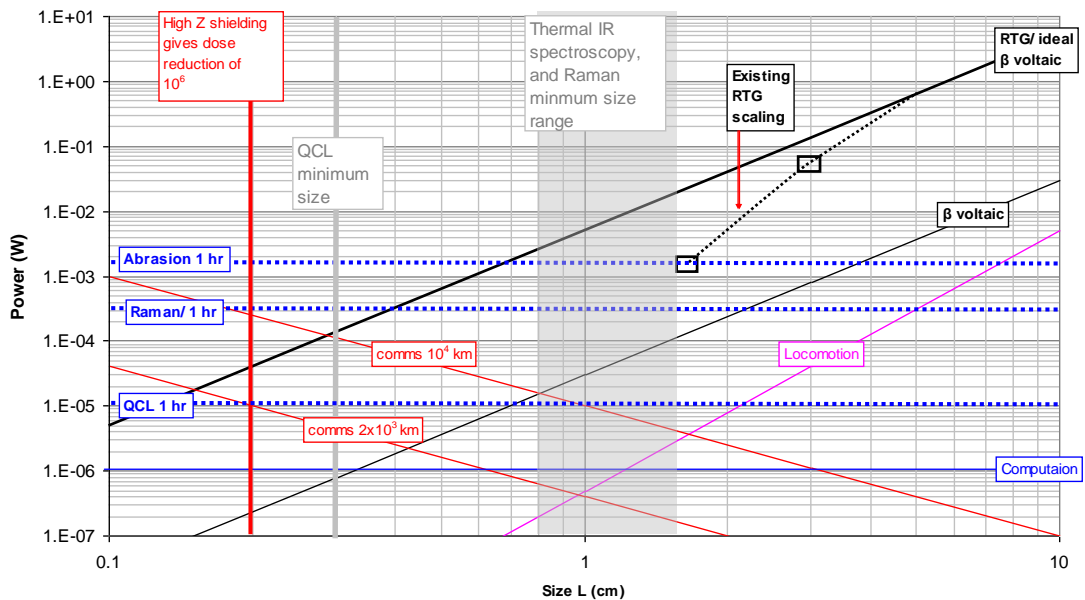

*Figure 4 shows the scaling of RTG powered rovers. The smallest RTG size currently described (~ a 1.5 cm/~4.3 cm$^3$ cube [101]), has sufficient power to provide all rover functions. Smaller sizes seem possible for vacuum insulated RTGs. The only problem would occur in Mercury dayside environments, where T environment is similar to T radiative equilibrium for a mm scale RTG. Communication problems begin at a scale size of 3 mm. Smaller α/βvoltaic devices are currently experimental only. Increasing limitations are seen for a rover powered by current betavoltaics below 3 cm, with communication limits becoming problematic below ~ 1 cm. Operating at betavoltaic theoretical energy density, the energy scaling would be close to the RTG 5 mWcm$^{-3}$ lines. Energy requirements for locomotion scale more rapidly, at L$^{-4}$, due to the reduction in absolute velocity with size. QCL lasers may be limited by capacitor size to 5 mm.*

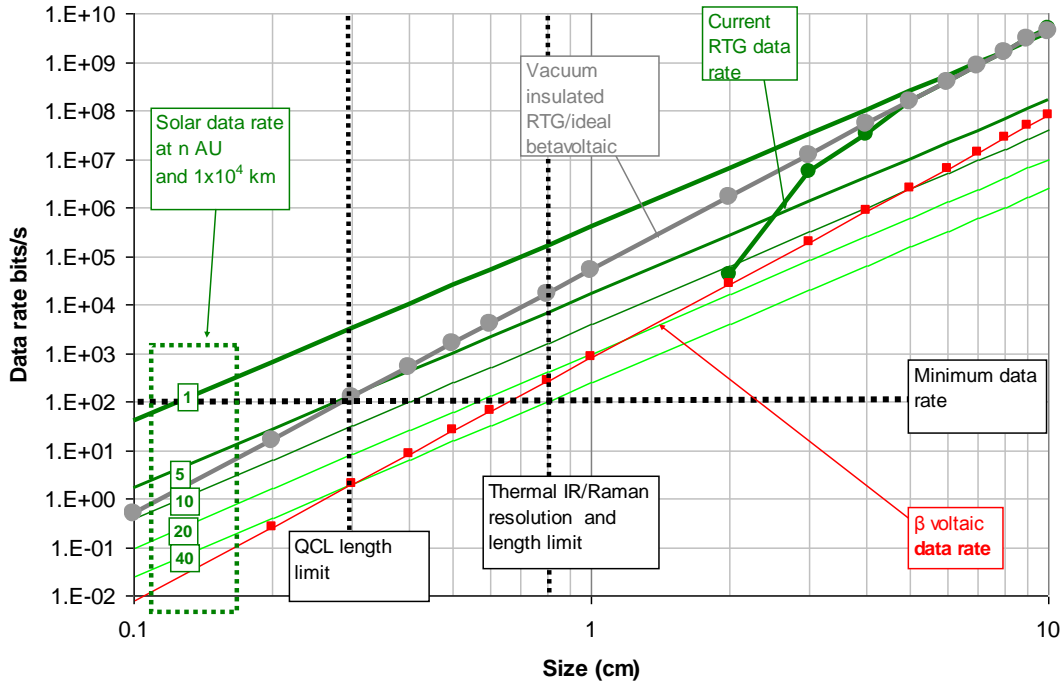

*Figure 5. The effect of rover scale on data rates. At large sizes RTGs are superior to solar on a size basis at > 1 AU. Current RTG size efficiency falls of to a minimum at the smallest current RTG size (~2 cm). Solar is good to the composition sensor size limit at ~5 mm to ~10 AU (Saturn), and at 10 mm to 40 AU (Pluto). Betavoltaics are inferior to solar and RTGs over much of the design space due to their low power density, though if operating at theoretical efficiency they would be good to 3 mm, as would vacuum insulated RTGs. At small scales due to the $L^3$ vs $L^2$ scaling of isotope sources relative to solar power, solar becomes less unfavourable at greater distances.*

Modern electrochemical cells achieve quite high energy densities (0.2-0.4 Whrg$^{-1}$/ 720-1440 Jg$^{-1}$), and might be considered as a primary energy source. Reasonable rover exploration ranges and lifetimes are achievable at the envisaged data rates– see Fig. 6. Below 1 cm in size, communications becomes the major limit on electrochemically powered vehicle life, and reasonable length missions become problematic below 5 mm.

Secondary energy storage is important both for providing peak power loads and, for solar, for allowing operation in periods of low light and darkness. Electrochemical batteries are the main secondary energy stores in current spacecraft. Current rechargeable cells have energy densities of 0.2 Whrg$^{-1}$/0.7 kJg$^{-1}$. Near future cells look likely to achieve 0.4 Whrg$^{-1}$/1.4 kJg$^{-1}$. Very small Li ion cells have been developed using thin film techniques, with cell areas <1 mm$^2$ and cell thicknesses of 15 μm [106]. Current primary Li based cells can achieve 0.4 Whrg$^{-1}$/1.4 kJg$^{-1}$. Fuel cells can achieve high energy densities (1.85 Whrg$^{-1}$/6.7 kJg$^{-1}$), but for the small scales being considered here, the need to store O$_2$ and H$_2$ is problematic. H$_2$O$_2$ could be used as a liquid phase O carrier at a mass penalty of ~2 (0.93 Whrg$^{-1}$/3.3 kJg$^{-1}$). N$_2$H$_4$ has been used directly as fuel, but raises the mass penalty to 2.7 (0.7 Whrg$^{-1}$/2.5 kJg$^{-1}$). Fuel cells have the potential to be scaled to very small sizes. The primary limitation of electrochemical cells is the need for a liquid (or solid) electrolyte with good ionic conductivity. This restricts battery operation to relatively high temperatures. Low operating temperatures for Li based cells are -40$^{\circ}$C, with the possible extension to -60$^{\circ}$C [107]. As far as I am aware, electrochemical cells using cryogenic electrolytes (e.g. semi-polar organics acting as ion transporters dissolved in cryogenic liquids) have not been investigated.

Capacitors can be used for energy storage, with electrochemical ultracapacitors having energy densities approaching Pb acid batteries at 0.01 Whrg$^{-1}$/36 Jg$^{-1}$. Ultracapacitors have the advantage of very rapid discharge and virtually unlimited cycle life. However they will probably be restricted by electrolyte cooling/freezing in the same way as more conventional batteries and fuel cells. Non-electrolytic capacitors can be used at much lower energy density – 0.1 mWhrg$^{-1}$/0.36 Jg$^{-1}$ [108], but power density is 5 Wg$^{-1}$. High dielectric constant materials operating at cryogenic temperatures (77 K) could boost volumetric energy densities by a factor of 10 [109]. Capacitors are scalable to sub-micrometer sizes.

Mechanical systems such as flywheels and springs are also candidates for low temperature energy storage. Flywheel energy density is ~ 0.05-0.1 Whrg$^{-1}$/180-360 Jg$^{-1}$ [97]. Spring-based systems have a lower energy density estimated at 0.1-1 mWhrg$^{-1}$/0.25-2.5 Jg$^{-1}$ (500 MPa operating stress, 50 GPa modulus composite, 10% system efficiency if rover structure is not used for bracing). More efficient elastomers are temperature limited to -100-+250$^{o}$C. Problems of flywheel based systems are gyroscopic effects if rotational speed is high, though this is not necessarily an issue if the system is used for peak power provision, or for housekeeping power during periods of darkness. Losses due to friction may become significant if power needs to be stored for long time periods, e.g. a lunar night. At a 10s of cm scale, assuming magnetic bearings and a vacuum enclosure, losses are ~ 0.1% per day [97]. Frictional losses will probably scale as L$^{-1}$, suggesting that power losses will be 1% a day at ~1 cm scale and 10 % a day at ~ 1 mm.

It should be noted that one option for solar powered rovers is not to operate at all in darkness, and thus avoid the energy storage problem. This was the strategy used by the Muses C microrover [8].

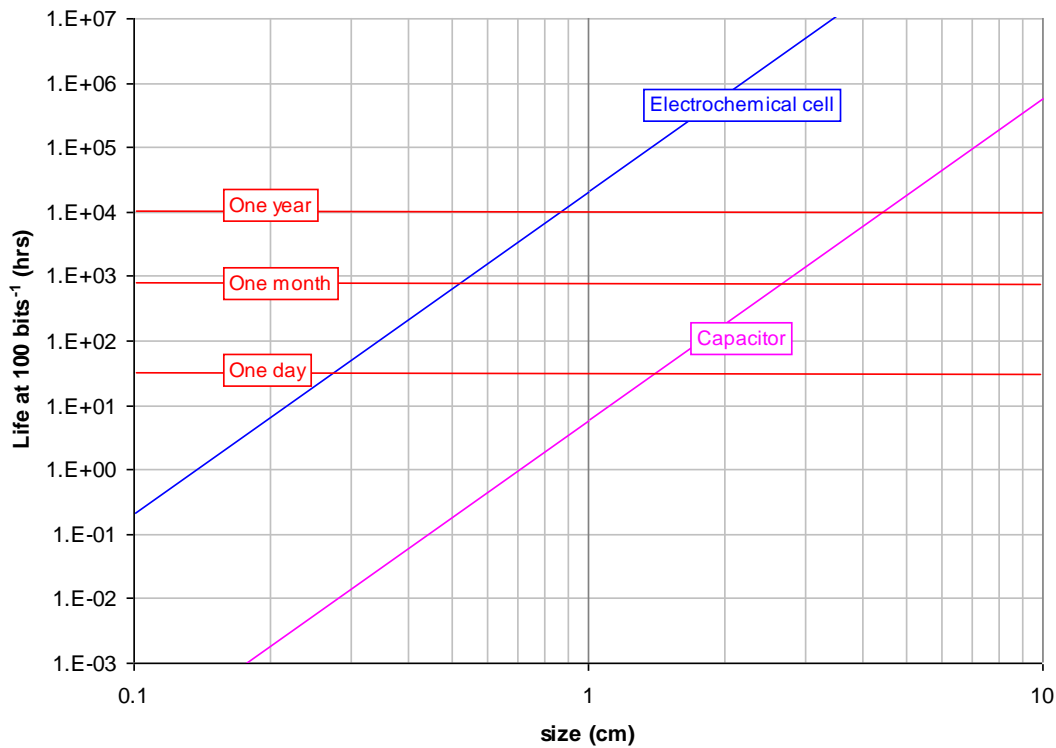

*Figure 6 shows the effect of scaling on battery and thin film capacitor life, based on transmitter power requirements of 10 µW at a 1 cm size scale, and cell/battery mass fractions of 50%. Rover life scales adversely with size due to L$^3$ scaling of energy stored and L$^2$ of antenna gain. A battery powered rover could send data for two years at a scale size of 1 cm, and one month at a scale size of 5.5 mm. For capacitors, rovers larger than a 1.5 cm size scale are needed to have transmission durations of more than a day, but due to L$^5$ scaling, a 4.5 cm scale rover could transmit for a year.*

Time taken by small sensors needs consideration when power limited, or when analysis time scales adversely with size. Figs. 7 and 8 show the effects scaling on analysis time for power sensitive Raman and QCL analysis, and areal flux sensitive sensors such as APX and natural γ. RTG powered rovers can perform these analyses in reasonable time at all currently relevant size scales (>2 cm). The low power density of betavoltaics is a major disadvantage for power dependent processes. However, vacuum insulated RTGs or ideal alpha/betavoltaics would allow operation of these energy intensive processes down to 6-10 mm. Raman and QCL analyses become solar limited at the minimum sensor size of about 5 mm at 5 AU, approximately Jupiter's orbit, though abrasion will be a time limiting factor at this distance. Larger solar powered rovers are more time efficient, and are comparable to the smallest currrent RTG rovers at distances less than 10-20 AU. Current APX becomes slow at ~1 cm. APX redesign (see section iv) will allow 3 mm scale sensors to perform as rapidly an as the current MSL APX (~15 min).

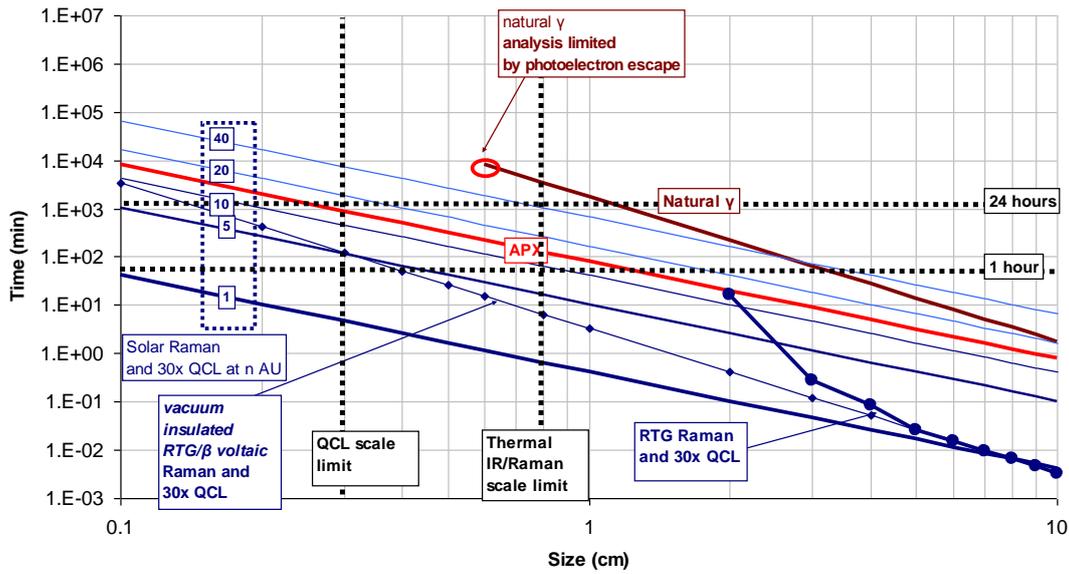

*Figure 7.* The scale effects on analysis time for power dependent instruments and for sensors dependent on low photon/particle flux. For power dependence solar is plotted from 1-40 AU. RTG powered scaling is also plotted. The size limit due to diffraction/fabrication for QCLs, Raman and thermal infrared spectroscopy is also shown. The Raman line is 30x the QCL time – separate QCL lines have been omitted for clarity.

The APX design scaling is based on present designs, and not the improved version discussed section iv. Abrasion data is not plotted for the purposes of clarity, but follows Raman and QCL scaling quite closely.

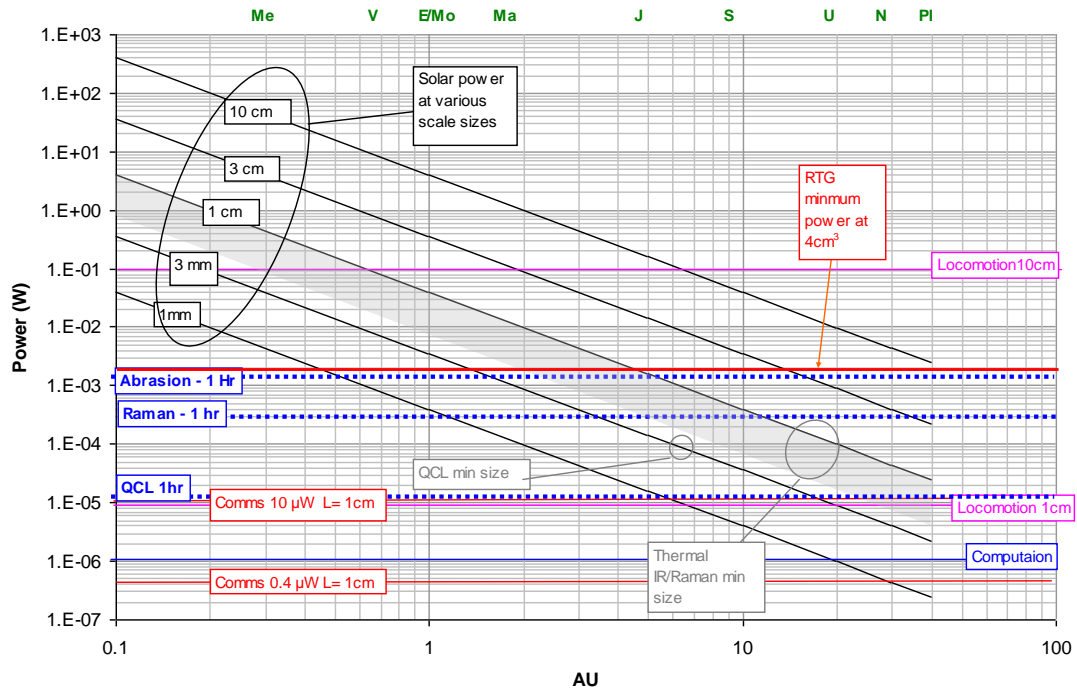

*Figure 8.* The effects of distance from the sun on the scaling of solar powered rovers. Locomotion at larger sizes, at the speeds envisaged, becomes a problem due to the square/cube effect of rover mass to solar panel surface area. This is the case for 10 cm scale length solar powered rovers at 20-30 AU. 1 cm scale experiences communication problems at ~100 AU. The minimum size to power abrasion and Raman/QCLs at 5-10 AU is 1 cm scale.

*The smallest current RTGs, at 2 cm, are fine for all tasks examined, despite the fall off in efficiency at small size. Vacuum insulated RTGs have not been shown for reasons of clarity, but should scale well to mm size. 1 cm scale communication energies are shown for comparative purposes. Size sensitive communication scaling is shown in figure 5.*

Finally rover actuators need to be considered. 0.1 mm diameter electrostatic and 1mm diameter piezoelectric motors have been built [110]. 2 mm diameter electromagnetic motors are available commercially [111]. Motors based on the giant magnetostrictive effect should show similar scaling to piezoelectric motors. Scaling has been reviewed in [111]. Electrostatic, piezoelectric and electromagnetic motors are felt to have similar fundamental scaling properties. Earlier, pessimistic, scaling assessments of electromagnetics did not consider permanent magnets, or increased current density due to improved heat dissipation at small sizes. Scalability of all rotary types into the immediate sub millimetre size range seems achievable.

Miniature linear piezoactuators are capable of very large bending strains (e.g. 150$^\circ$), and require very little power and relatively low voltages (15V) [112]. These could be used as leg, deployable sensors and pointable antenna actuators.

Actuator scaling does not appear to be a fundamental problem in scaling in the centimetre to millimetre range.

*vi) Thermal environments*

*Small rovers have a large surface area to volume ratio and the operating temperature of key components will be close to that of the environment, with the exception of some radioisotope powered designs. Above 300$^o$C non-Si, AlP, GaP, AlAs and alloys, and wide band semiconductors (e.g SiC, GaN, AlN, diamond) are likely to be needed. The primary target where they are essential is the Venusian surface. The Al/Ga-P/As system should cover the Venusian temperature range, and technologies in these materials are well developed from LEDs and GaAs heterojunction transistors. Analogue thin body SOI operating has been successfully operated at 400$^o$C as have a variety of GaAs devices and circuits. Due to the maturity and high integration densities of these two semiconductors, a detailed study needs to be made of their suitability for operation as dense integrated circuits at 390-480$^o$C. Wide band gap semiconductors are far less mature than Si, but developments in the field suggest that dense integrated circuits operating at Venus surface temperatures should be feasible in the near future. If these match the performance of 1-2 μm GaAs in the 1980s, computational power dissipation for a Rocky-style rover would be 0.2 W. Subsumption-style designs can be built with very low transistor counts, and could use less advanced high band gap circuitry, which would allow smaller semi-autonomous rovers to be built for Venusian exploration. Wide bandgap energy dispersive sensor materials need to be developed for 390-480$^o$C operation. Diamond based semiconductors may be viable to ~1300$^o$C, which would be useful for deep gas giant exploration.*

*With some design changes, modern CMOS can operate down to deep cryogenic temperatures. Cryogenic temperatures allow more sensitive detector materials and architectures to be used. As discussed in section v, the key challenge for microrovers operating at cryogenic temperatures is secondary energy storage. Thin film capacitors are a viable option. Propellant storage may become an issue for small landers, especially when $O_2$, $F_2$ and $CH_4$ freeze. Solid propellants and cold gas $H_2$/He based cold gas propellants could be used instead.*

Small rovers have low thermal inertias and, with possible exception of larger radioisotope powered rovers, will run at ambient temperatures. In biological terms they will be cold blooded. At a small vehicle size, assuming radiative losses only, and ~300 K temperatures, times to thermal equilibration will be 100s of seconds at a cm scale, and tens of seconds at a mm scale. Polished low emissivity surfaces are assumed not to be an option for rovers, due to dust contamination.

The temperature environment, and its effect on rover systems, is summarised table 6 and Fig. 9.

The temperatures that may have to be designed for range from up to 480$^o$C on Venus or equatorial dayside Mercury, to less than 50 K in the outer solar system, and closer to home, in lunar polar craters. Sedna, at apehelion of ~900 AU would have a noon surface temperature of ~12 K, though it is now closer to perihelion at about 90 AU, and noon temperatures are in the region of 30 K. In the case of Mercury, high temperatures can be minimised, to some extent, by exploring only high latitudes during daytime.

| T °C | T K | Comments |
|---|---|---|
| 480 | 753 | Venus surface*/Dayside equatorial Mercury |
| 400 | 673 | Extended Si electronics range |
| 300 | 573 | Max T Si electronics |
| 250 | 473 | Max T elastomers |
| 120 | 393 | Lunar dayside |
| -40 | 233 | Min T current Li ion cells |
| -60 | 213 | Possible extended low T Li ion |
| -100 | 173 | Elastomers stiffen |
| -173 | 100 | Lunar/Mercury night |
| -203 | 70 | All reactive propellants except $H_2$ freeze |

*Table 6. The temperature range of some solar system exploration targets, and ranges of operation of various spacecraft technologies.*
*\*Venusian surface temperature is altitude dependent and ranges from 390°C at Maxwell Montes (+10.4 km) to 485°C at Diana Chasma (-2.9 km). Temperature at average elevation is 460°C [113]. All would be favoured targets as the crustal composition is likely to vary from 'continental crust-like' in the highlands to denser 'oceanic crust-like' in the topographic lows.*

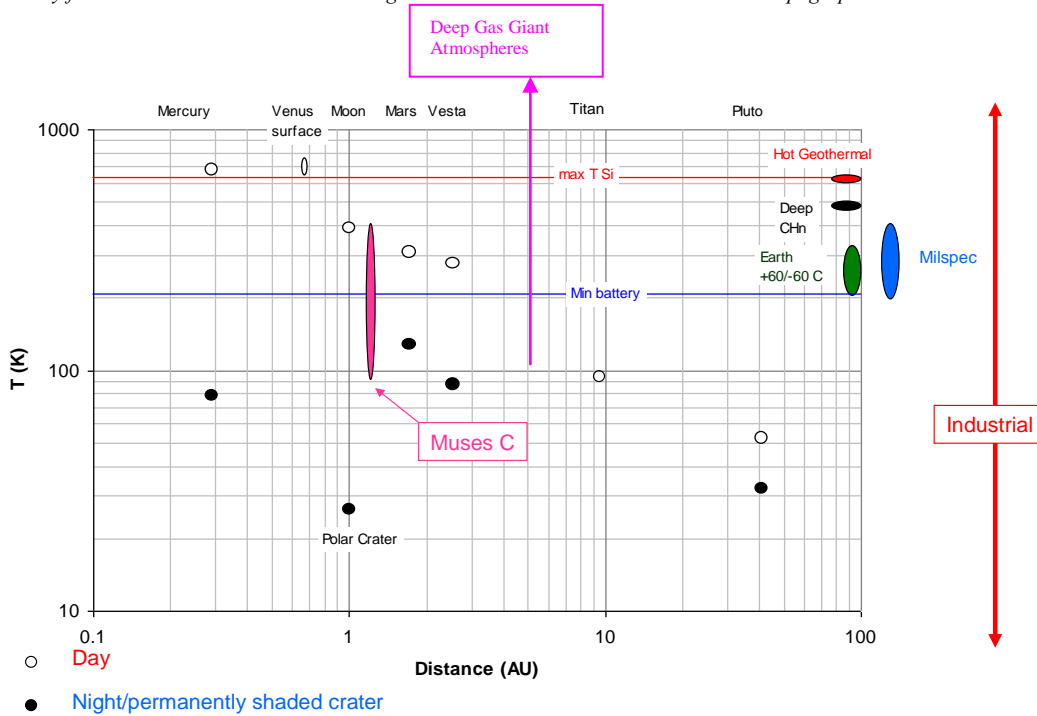

*Figure 9 shows the temperature regimens likely to be encountered in solar system exploration, and comparable terrestrial thermal environments, together with the widest small rover design temperature range to date (Muses C).*
*Venus surface shows minimal diurnal temperature variation due to the thick insulating atmosphere.*

The problems of very high temperatures primarily stem from the low band gap of Si. At temperatures greater than 300 °C (perhaps as much as 400 °C - [114]) leakage currents from thermally excited charge carriers greatly complicate the effective operation of even SOI transistors. GaAs is also generally used at lower temperatures, but has a higher band gap (1.4 vs 1.1 eV). Work by [115] demonstrated operation a variety of digital and analogue devices at 400 °C, with some operating at 500 °C.

Higher band gap, less developed semiconductors will probably need to be used, though thin body SOI, and GaAs should be investigated for possible Venus surface electronics, especially at the lower end of the surface temperature range. AlP, AlAs, GaP and alloys have band gaps which should allow operation at Venus surface temperatures, and have a relatively mature technology base from LEDs and GaAs heterojunction transistors [116]. The more high temperature capable, wide band gap candidates include SiC, GaN, and AlGaN. These semiconductors typically are much less mature than Si/GaAs, and historical integration densities have been low. Additionally the primary type of transistor for these semiconductors is a single carrier type MESFET/JFET, which in its simplest form, has a high power dissipation per device, potentially greatly limiting device densities. Unlike a complementary logic such as CMOS, these transistors draw power continuously when not switching, and this raises their power delay product, especially for systems with low switching activity. Complementary architectures require both n and p type transistors, and are most efficient in a semiconductor like silicon with similar hole and electron mobilities. For low temperature applications of semiconductors such as GaAs, where electron mobility is much greater than hole mobility, CMOS like architectures have not been aggressively developed, since the low hole mobility will result in low operating frequencies similar to, or worse than, better developed ambient temperature silicon. The single carrier power issues have been overcome using pseudocomplementary logic styles in ambient temperature GaAs (e.g. [117,118]), or for low duty systems, by powering down when not active. These techniques should be applicable to the high temperature single carrier semiconductors.

Much work is being done on developing high band gap semiconductors for >300°C use, as applications are likely in monitoring aerospace and industrial gas turbines. GaN and SiC also have uses in power electronics where heat dissipation, and hence operating temperature is potentially high. Major drivers for GaN and AlN device miniaturisation also come from their use in optoelectronics and high frequency analogue electronics. There is also a major program to develop deep submicrometer (10-20 nm) III/V on silicon electronics, where they have potential advantages over more conventional silicon logic (e.g. [119,120]). CVD diamond is also being actively researched, as it has properties which suggest extremely high performance for a variety of electronic devices. Recently high temperature interconnect problems on SiC chips have been partly overcome at 500 °C [121]. Some of the problem appears to arise from poor interconnect oxygen resistance [121], and could be resolved by sealing devices in an inert atmosphere. Operation has been demonstrated at 500 °C, in air, for $2x10^3$ hrs. Device densities appear to be $5x10^{2-3}$ cm$^{-2}$ for logic, and $5x10^4$ cm$^{-2}$ for memory, though at present only mm scale chips have been made using a 10 μm design. The authours plan to build chips with $10^2$-$10^3$ devices in the near future using a 6 μm design rule [122]. Scaling to mid 1980s GaAs device densities is felt to be readily achievable [123]. In the case of systems based on In/Al/Ga nitrides, short term operation (~24 hrs) of transistors with minimum features of ~0.25 μm has been demonstrated at up to 1000°C in vacuum [124]. The Arrhenius nature of device degradation rate with temperature suggests that, at T <700-800 °C, much longer device lives should be feasible. Small circuits, for example an integrated differential amplifier, have been demonstrated at 500 °C [125]. The size of the devices in [124] implies device densities of $10^2$-$10^3$/cm$^2$ for logic and $10^4$/cm$^2$ for memory.

At the time of writing, 500°C circuits of 10s to 100s of devices and a few mm$^2$ are available. However it seems likely that developments in the field will proceed rapidly and integration densities of up to $10^6$ devices or more per cm$^2$ should be achievable in the near future for SiC and III-V nitrides, with long term operation at ambient Venusian surface temperature. By analogy with ambient temperature GaAs memory (e.g. 4 K SRAMs designed for the Advanced On-Board Gallium Arsenide Signal Processor, with GaAs selected primarily on the basis of radiation hardness, [126]) such devices are likely to dissipate ~ 1 W/Mbit when aggressively designed for low power. A simple 8 bit processor could be built with ~4-6000 transistors (e.g. the Motorola 6800 or the Intel 8080), and based on results with 2 μm GaAs technology (~1 W for a 11,000 transistor, 16 bit multiplier, [127]), would dissipate about 0.3 W. This suggests that a Rocky III style rover would have a computational power dissipation of 0.5 W.

Some of the wide band gap semiconductors are being developed for high frequency operation, and in the case of Venus, the main control on frequency use would be atmospheric transparency of the band concerned.

The very low transistor counts of subsumption style control systems suggests that revisiting such architectures may be beneficial. If compiled into silicon, Tooth or Rocky III style rovers would need 1600-5000 transistors operating at 10s to 100s of Hz (see section i). Power dissipation would be low, especially if a pseudocomplementary, or CMOS-like architecture were to be used, and mW power levels could be anticipated. Complementary JFETs compatible with modern SiC wafer quality have been modelled in detail. Static powers are at the 0.4 μW/gate level using a 2 μm rule for -50 - +600 °C [128]. This suggests a 0.15-0.5 mW static power dissipation is possible for Rocky III style rovers.

In the longer term, assuming all the technological problems of building low power, >300°C, dense, silicon CMOS equivalent logic are overcome, fundamental power dissipation in computing is limited by the need to overcome thermal noise, so a 760 K circuit on Venus will dissipate ~2.5x the power of a 300 K circuit on Earth.

For higher temperatures, e.g. deep gas giant atmospheric probes, extrapolation of maximum temperature of electronic operation from band gaps ceases to be straightforward. This is because of a variety of effects. At more than half melting temperature dopants become mobile, and reactions occur between metallization and semiconductors, although the latter can be mitigated by the use of appropriate barrier layers. This will impose limits on some Ga and Al based systems of 600-700°C [116]. Diamond substrates based on CVD single crystal diamond substrates show very low diffusion of dopants to 1500 °C [129], but will be limited to about 1300 °C by transformation to graphite. BN based devices will have similar thermal limits partly due to dopant diffusion, and, if cubic BN is used, graphitisation. Microvacuum electronics (with a 10x10x10 µm device size) does not offer a solution. Here the limitation is due to vacuum volatisation of metals from conductors, ionic conductivity problems and vacuum volatisation of oxygen or nitrogen from refractory insulators. The latter will probably restrict use to 700-800 °C. However, radiation resistance is likely to be extremely high.

Energy dispersive sensors operating at 300 K need a bandgap of 1.4 eV [130] to prevent thermally excited charge carriers interfering with signal induced charge carriers. Energy dispersive scintillators can be used, as can higher band gap III/V semiconductors (e.g. AlN 6.5 eV). These should be able to operate at 460°C where a bandgap >3.3 eV is required. A variety of oxides are currently being developed for use in transparent transistors – where bandgaps >3 eV are needed [131]. These materials require much work, as they have not been used as radiation detectors in the past. Efficiency for MeV γ rays, e.g. from $^{40}$K, is not a problem at the Venusian surface. U/Th concentrations can be determined from spectroscopic analysis of low energy decay chain gammas, allowing K concentration to be determined from the non U/Th background (GCR gammas are negligible due to atmospheric shielding). At this temperature, energy resolution may be reduced by a factor of 2-3 compared to a 300 K lower bandgap detector. Scintillator based designs may be feasible. Deep hydrocarbon exploration instruments can operate at 220°C and are based on a rare earth doped perovskite scintillator, and SiC avalanche photodiodes [132]. Similar systems should be extendable to Venusian conditions.

Based on the scaling of band gaps of infrared photodetectors used at 300 K (e.g. InSb 0.17 eV – [133]), semiconductors with band gaps > 0.43 eV should be usable as ambient Venusian temperature photodetectors. An alternative strategy would be to use thermistor or bolometer infrared vision systems, similar to those used in ambient temperature (300 K) thermal imaging systems [133]. These would utilise emitted and reflected infrared light, and exploit differences in surface emisivity and sky illumination effects. Slow, mechanical, image dissecting systems would allow use of single/small thermal detector arrays, which may fit better with current low device integration levels.

Battery systems are not so much of a problem, as a range of high energy density, high operating temperature cells have been investigated for electric vehicle use, and are used in pyrobatteries (e.g. the $FeS_n$-LiX systems). A number are suitable for use under Venusian conditions, and the pyrobattery operating temperature is the same as the Venus surface temperature range (e.g. [113, 134]). Dry lubricants are available for high temperature use, and electric motors have also been designed for operations at 540°C [135]. Potential piezolectrics (e.g. $La_3Ga_5SiO_{14}$) of interest for actuators maintain their electrical properties to 750°C [135].

Thermal generation of hole-electron pairs will reduce the thermal efficiency of PV cells in a high temperature environment. The effect is greatest for lower band gap semiconductors such as Ge and Si. Higher band gap semiconductors can be used, but these suffer an efficiency loss due to a poor match of their absorption with the solar spectrum. For example, CdTe efficiency will fall from 30% at 0°C to 8% at 300°C. A semiconductor with a higher bandgap of 2.8 eV will have a reduced 0°C efficiency of ~12%, but will only fall to ~8% efficient at 300°C [136]. The efficiency fall will be compensated, in part, by the much higher solar flux available in the innermost solar system.

Despite the thick cloud layer, PVs are a possible power source for the Venusian surface [137]. Unfortunately their low power of ~1 $Wm^{-2}$ results in insufficient power for probable near term semiautonomous electronics, though as noted earlier, this is an area where change may occur rapidly. The high temperatures of Venus are not an overwhelming problem for RTGs, as hot ends operate above ambient Venusian surface temperatures. Efficiency will be reduced by the higher 'cold end' temperature. Given the low power of photovoltaics at the Venusian surface, RTGs will probably be the system of choice for long term surface exploration, unless a subsumption style architecture proves useful. These scaling effects are shown in Fig. 10.

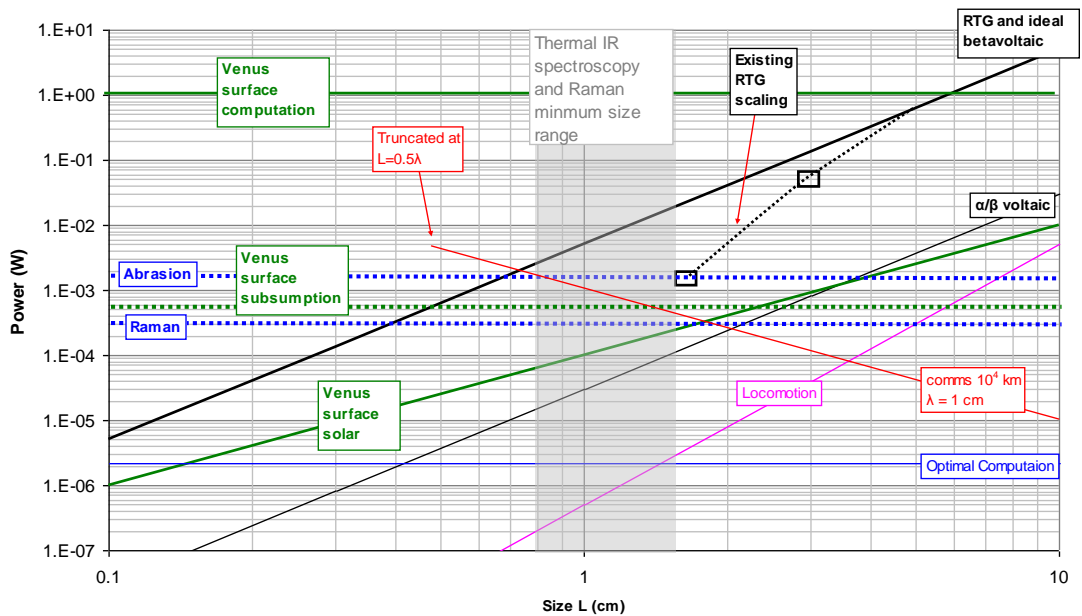

*Figure 10* shows scaling of various Venus rover subsystems. The major limit appears to be for computation assuming current/near future high temperature electronics are used. This sets a size limit of >6 cm for an RTG powered Rover. Smaller RTGs will not scale well, due to insulating difficulties in a hot high pressure environment. However, if beta or alphavoltaics operating at close to their theoretical efficiency are developed, and a hardwired subsumption based architecture is used, the computational requirements could be substantially reduced. The minimum size would fall to ~ 1 cm based on communication constraints through the Venusian atmosphere (Modelled absorption through a Venusian model $CO_2/SO_2$ atmosphere increases rapidly above 20 GHz [138]). The communication plot is therefore terminated at 5 mm, ~ 0.5 λ for 30 GHz. Optimal computation refers to high temperature electronics with a power efficiency similar to current low power CMOS. Solar power appears viable for 2-3 cm scale rovers, and is limited by computation and ultimately by communications. QCLs are omitted as operation has not been demonstrated in the Venus surface temperature range.

Low temperatures present a different set of problems, but also some opportunities. The key problem area is that of low temperature secondary energy storage (see also section v). An extra problem arises if there is cycling into a low temperature zone of operation where electrolytes can freeze and experience severe volume changes, which lead to cell failure on rewarming. This can be overcome with solid state electrolytes (e.g. freeze tolerant Li cells in [106,139]) or with simple battery redesign – freeze tolerant Pb-acid batteries are commercially available (e.g. [140]). Terrestrial high temperature battery systems developed for high energy density, which are solid at room temperature, are often able to withstand repeated freeze thaw cycling (e.g. [141,142]). The ability of these systems to withstand freeze-thaw bodes well for redesigning other conventional cell chemistries to tolerate this challenge. The high temperature sulphide systems tolerance for freeze thaw is also useful for cells that are entering a Venus, or dayside Mercury hot environment, from a cooler transit regime. An alternative strategy is to have no secondary energy storage and simply operate only when illuminated.

The variety of usable semiconductor detector materials rises, as lower band gap materials become viable as temperatures fall. Examples include the HgCdTe system which has excellent sensitivity in the 5-10 μm range at ~70 K [133]. Band gaps rise slightly with falling temperatures, but exceptions occur, (e.g. PbS [133]). Design of mechanical systems should be reasonably straight forward, as there is considerable terrestrial experience in cryogenic bearing operation.

Low temperatures also pose problems for rocket propulsion systems. Conventional rocket fuels (e.g. liquid $O_2$ and $CH_4$) are space storable from Mercury to Pluto [143]. Propellant freezing may be an issue for small spacecraft, especially landers and sample return vehicles, though these can make use of low emissivity insulation. If a suitable liquid propellant combination is not available, a solid propellant rocket designed to give approximately the landing delta V may be needed. The solid propellant may also need to include additional fibrous or particulate fillers as crack arrestors to counter low temperature brittleness. Cold gas thrusters would be used for attitude correction, and also for final delta V adjustment. Unfortunately most of the mass efficient gases will freeze at outer solar system temperatures leading to the use of neon and then less mass efficient hydrogen and helium. (Helium and hydrogen have high exhaust velocities – but this is offset by the high mass fraction of the pressurised tank).

As minimum energy per bit operation is directly proportional to temperature, logic power requirements will fall in proportion to temperature e.g. 1/10 of 300 K operating energy at 30 K in the outer solar system. Deep cryogenic temperatures of, for Si <30 K, are associated with a 'freeze out' of dopant atoms. This is not a problem for current CMOS technologies, and CMOS logic has been designed for operation at liquid helium temperatures of 4.2 K and below (e.g. [144]).

*vii) Radiation*

*Modern CMOS SOI technologies are highly radiation tolerant, and suitable for most space environments using dual redundancy and hardened voting circuitry. Very high dose rates are a serious problem for exploration of the inner Galilean satellites, but appropriately designed CMOS is suitable for typical mission durations. The dose rate may cause issues for PV systems. The high radiation belt intensity encountered at the inner Galilean moons would potentially allow surface XRF measurements, but much work would need to be done on detector scaling, fields of view and shielding.*

The space radiation environment poses a number of problems, primarily for PV and electronic systems. Radiation has a total dose effect, where an accumulation of charge carriers trapped in electrically active insulators and displacement damage to bulk semiconductors can cause system failure. Additionally, high energy charged particles can result in enough energy deposition to switch the state of a device, or for some technologies, cause permanent electrical damage.

Total dose effects used to be a serious issue with CMOS technologies. CMOS depends on electrically active thin oxide films, which could accumulate radiation induced charge carriers. However as device size has reduced, the thickness of these oxide layers has fallen, and charge carriers can now migrate or tunnel out. Modern CMOS can tolerate total doses of in excess $10^5$ Grays [40]. Scaling also has beneficial effects on single event tolerance. While charges representing device state are much smaller, high energy particles deposit charge along a track, so a small device has less volume to collect charge. Additionally lower voltages are used to boost energy efficiency, resulting in reduced radiation induced charge carrier acceleration. To 2010 (~50 nm scale), overall total dose hardness has risen. For both SRAMs and DRAMs, SEU rates have been stable, or in the latter case, have fallen [145]. One potential problem at extremely small scales, probably less than 20-30 nm minimum feature size, is local single event radiation damage, where the damaged zone from a particle track is of the same order of magnitude as the electrically active portion of the device. This will be particularly important for heavy cosmic ray nuclei such as C and Fe. Another feature which boosts overall radiation hardness, is the use of electrically isolated semiconductor on insulator, with the best developed example being SOI. This is being adopted to mitigate electronic issues with device scaling in standard Si, and is likely to become a major terrestrial semiconductor technology at small device sizes. Semiconductors in PVs are also susceptible to radiation and can tolerate $10^6$ Gray of 10 MeV electrons with 50% power loss [136]. Proton dose tolerance is ~100 times lower.

Radiation environments and types, and their problems are summarised next. GCRs are primarily high energy protons, but with heavier nuclei included in roughly stellar abundances. They cause single event upsets, but due to energies of GeV per nucleon, require a metre of more of water equivalent shielding. The fluence of these is low, and can be coped with by error detection and redundancy, and selection of device technologies which are immune to single event hard errors. A second problematic area is a solar particle event, which can generate large numbers of energetic protons. These are shortlived events, but particle energies are high, and doses potentially on the order of 30 Gray for a large event. A small event will cause increased single event errors. A very large event may require temporary shutdown and the presence of a small amount of highly single event tolerant 'housekeeping' hardware. However such large events are infrequent.

The final, most challenging area, is that of the magnetically trapped particle radiation belts. Here I will concentrate on the gas giant belts and Jupiter in particular. It should be noted though, that Earth's inner Van Allen proton belts boosts particle energies of more than 400 MeV, and place strong constraints on satellite orbits. In the Jovian system the problematic particles are predominantly high energy electrons. These rapidly give high total doses, especially on the inner moons, Europa and Io. On the Europan surface, with 1 mm H$_2$O equivalent shielding, the 30 day dose is on the order of $2 \times 10^4$ Grays. This can be mitigated to some extent by shielding. 1 cm H$_2$O equivalent gives a dose of $3 \times 10^3$ Grays over the same time period, while 10 cm gives a total dose of 80 Grays [146]. High density shielding material, e.g. W, will reduce dose by a factor of 10-20 for an equivalent thickness, and would give a Europa surface dose of $3 \times 10^3$ Grays for ~1 mm of high density material. This level of shielding would allow long lived missions to the surface of Ganymede, Europa and Io. Imaging sensors would be shielded in part by high Z, high density, optical materials for example BGO. However, CMOS arrays with very high total dose tolerances (to $10^6$ Gray) have been designed [147].

Detailed analyses of the Europa radiation environment are available in the literature, though far less is available for Io. [148,149] have examined fast, elliptical orbit, flybys and found tolerable doses. Current models suggest a 5-10 times increase in dose relative to Europa [150] (see Fig. 11). This may be overly pessimistic, as the Galileo CCD imager noise only increased by 20% in Io orbit relative to Europa orbit [150]. As discussed earlier, $10^6$ Grays electron dose is a reasonable limit for hardened electronics and for PVs. This suggests about 0.2 gcm$^{-2}$ (2 kgm$^{-2}$ for solar panels) seems the minimum to allow for a 1000 day Europa mission, and if extrapolation from Galileo is reliable, similar shielding should be sufficient for Io. Using the more pessimistic modelled estimates, 2 gcm$^{-2}$ shielding would be needed for Io rovers/orbiters. This would increase the mass of orbiter solar panels and 1 cm scale rovers by a factor of 10-20. The relative mass penalty would be smaller for larger rovers, being about 200 g for a solar powered 10 cm scale rover. For orbiters this weight penalty could be avoided by a concentrator design, with only a small area of photovoltaics to shield [151]. Radioisotope systems are likely to be relatively long lived in inner Galilean environments due to shielding by bulk structure of the more sensitive components, but issues may arise with inadequate shielding for the smallest RTGs discussed, as well as even smaller scale betavoltaics.

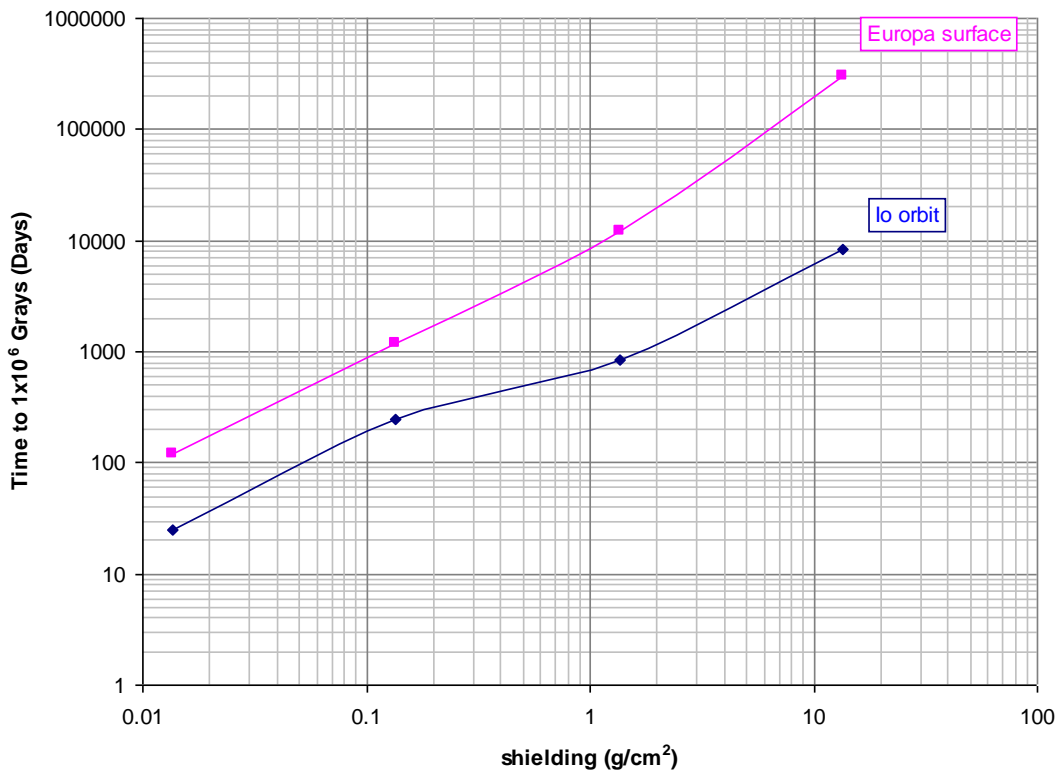

*Figure 11 shows the effect of shielding on time to a radiation dose of $10^6$ Grays, which is a reasonable limit for PVs or hardened electronic systems. The Europa shielding is based on [152], the Io on [148,150], but the real Io surface/orbit dose may be closer to Europa. 1000 days probably represents a minimum mission duration, to allow for achieving a stable Io orbit, the deployment of several rovers in diverse environments and long term (orbiter based) surface imaging. Note that even a small amount of shielding (<<1 g/cm$^2$) has a significant effect on total dose.*

Lower energy components of the Jovian radiation belts may be a problem for optics. The total dose is much higher than the more penetrating radiation discussed earlier (see fig. 11). This radiation is easily shielded by 1 mm $H_2O$ equivalent, and typical small rover structural materials. The very high surface dose may damage optical surfaces, and it may be necessary to utilise movable protective covers.

The high energy electron flux impacting the surface of the inner Galilean satellites may allow operation of an XR detector in an XRF mode – the excitation of surface atoms being due to the MeV electrons. The detector would operate in the 1-100 keV region and a small, <1 mm$^3$, high Z detector would operate efficiently. The main problem with detector operation might be saturation with XRF photons and continua, and with high energy electrons. A detailed evaluation of shielding, detector speed, area and field of view would be needed to assess whether XRF would be feasible on Europa or Io.

*viii) Orbiters*

*Low mass, ~ 4 kg, orbiters could be designed for use to 10 AU. Further out, orbiter mass increases, with RTG power being most mass efficient. 20 kg orbiters are suitable out to Pluto. A mission to a Sednoid at an aphelion of 500 AU would require a 300 kg orbiter. RTG designs for communications power would also prove useful for powering a high exhaust velocity ion drive. PVs would have a similar role in the inner solar system.*

A key component of the exploration strategy outlined above is the relay orbiter. To set the orbiter scale an assumption has been made of a 1 m diameter, ~1 kg (mass based on [153]), rover data receiver dish. The Earth communications dish is assumed to be the same size. For solar powered orbiters the array area is also assumed to be 1 m$^2$, with a mass of 1 kg, and using modern advanced photovoltaics, would produce ~400 W at 1 AU. For a radioisotope powered rover, the mass of the RTG is assumed to be 1 kg, and would, using current thermoelectrics, deliver 5 W. 50% transmitter efficiency is assumed, and the receiving antenna is a 30 m deep space network class antenna. The maximum frequency for orbital relay to Earth communications is assumed to be Ka (30 GHz), as atmospheric absorption reduces the efficiency of higher frequencies. At 1 AU distance, this allows a data transmission of 25 kbits/s/$W_{electrical}$ for a 1 m orbiter transmitter sending data to Earth. An extra 1 kg of mass is assumed for additional subsystems. The orbiter data transmission rate is shown in Fig.12.

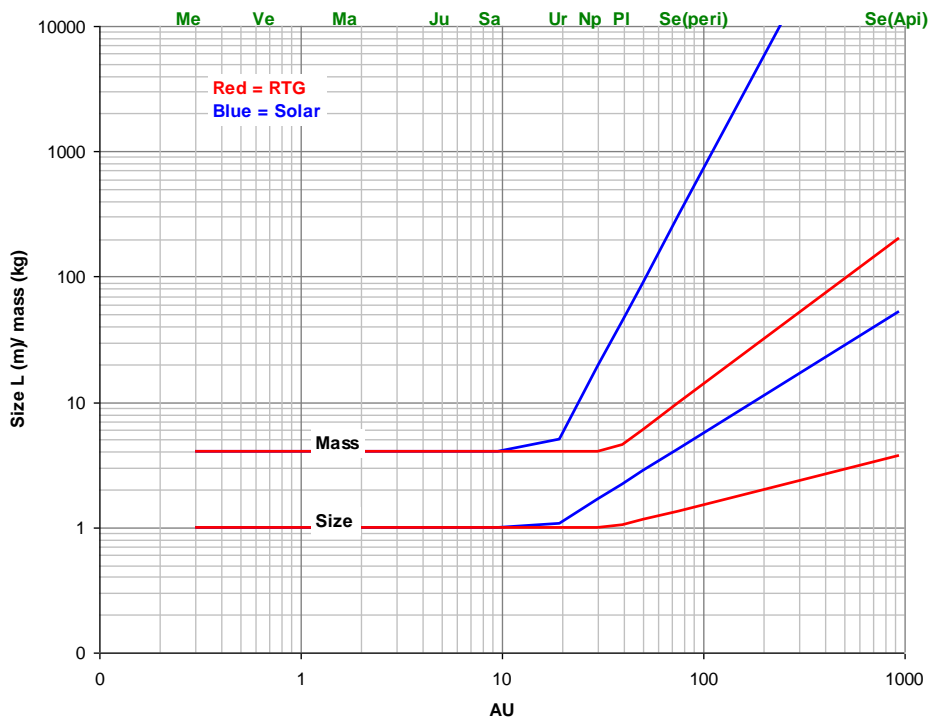

*Figure 12 shows the effect on orbiter size and mass on increasing distance from Earth. Dimensionally scaled data from [42] is used. Out to ~10 AU, mass is constant for a 100 bit/s minimum data rate. The mass is set by the need for a 1 m dish for rover – orbiter communications. (The kink in the graphs at 10 AU/30 AU is a plotting artefact). Further out mass increases rapidly for solar powered orbiters, but more slowly for radioisotope powered orbiters, which remain viable to 1000 AU+.*

At 1.25 Wkg$^{-1}$ the RTG allows an annual delta V of 0.5 kms$^{-1}$. For inner solar system photovoltaics, once power levels rise to 100 W, a delta V of 5 kms$^{-1}$ is feasible (ion drive efficiency of 50% and an exhaust velocity of 50 kms$^{-1}$).

*ix) Landing*

*Automated microlanders are feasible. Solid fuel/cold gas propulsion may be the best option in the outer solar system. Precision landing is possible with a relatively low computational load, when the target is on a vacuum world and solar illumination of target features is predictable. Aerobreaking is usable on targets with an atmosphere, and, where possible, is the more mass efficient option. However precise landing is subject to difficulties in modelling planetary atmospheric densities and/or windspeeds.*

Rover landing strategies will depend, to some extent, on the mass of the body being visited, but a more important factor will be the presence or absence of an atmosphere. A combined orbiter-rover pod can be envisaged entering a high orbit around the target (see Fig. 13). The rover pod contains multiple rover-lander pairs, and possibly other immobile probes such as penetrators and seismometers. The interior of the rover pod may be thermally controlled to allow the use of liquid fuels. Sub mm nozzles have been demonstrated (e.g. [154]). The rover pod will separate from the orbiter and drop to a very low orbit, from which it will dispense rover lander pairs. For a body with an atmosphere, the rover and lander will be in an ablative aeroshell, and after atmospheric entry and braking, the aeroshell will separate and parachutes will be deployed. The small size of the probes will allow the use of conventional parachutes on Mars, despite the thin atmosphere, which necessitates retro-rockets for larger landers (e.g. MSL). For a target with no atmosphere, deceleration from orbit will be by a solid or liquid fuelled rocket with a specific impulse in the 250-350 s range. Cold gas thrusters will be used for attitude control, and also final velocity adjustment and landing if main solid propellant rocket is used. For the vacuum targets being considered, acceleration due to gravity is very low and cold gas systems should not cause too much of a mass penalty. Selection of landing site could use an optical system. For vacuum worlds of known topography illumination can be accurately predicted, and simple thresholding of images will provide targets for tracking with an estimated 100 operations per pixel, and at $10^4$ binary pixels per map, 3 images increasing in resolution by a factor of 20 should allow landing accuracy of ~10 m starting from 100x100 km. Memory requirement will be on the order of $3x10^4$ bits. Similar strategies can be adopted for manoeuvring atmospheric entry vehicles, though the uncertainties in weather mean that more sophisticated image processing and/or more visibly outstanding landmarks would need to be selected. Additionally uncertainties in atmospheric densities will produce larger cross range errors that will need to be corrected. Other computational requirements will be low by modern standards (see table 1).

Based on previous lander experience, rover mass has been on the same order as lander mass, and for aerobreaked landings on Mars, the aeorshell/heatshield mass is equal to lander and rover mass [155].

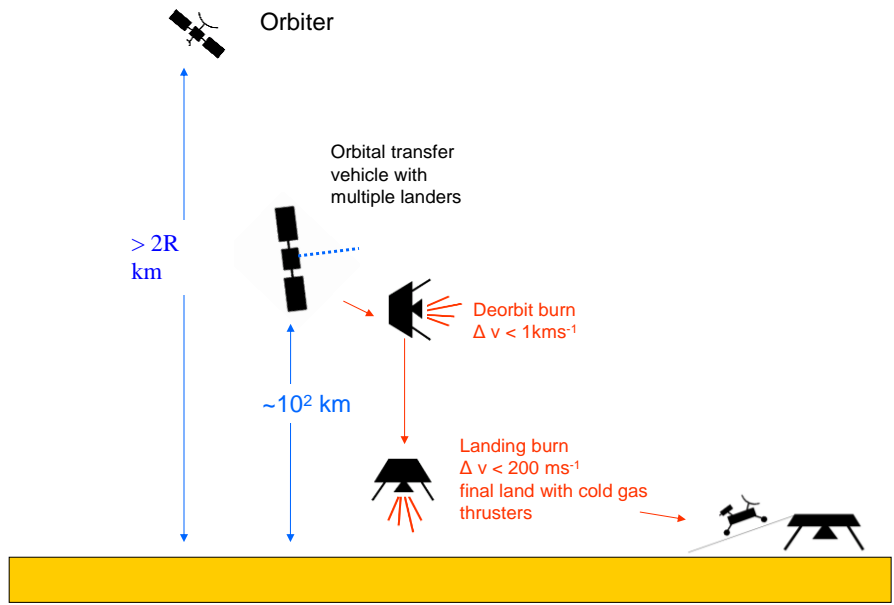

*Figure 13 shows the rover deployment strategy for a mission to a low gravity vacuum target.*

*x) Manufacture*

*10 mm scale microrovers are of the size scale of present day watch movements and small electric motors. Construction should be reasonably straightforward. Mm scale rovers have no close modern day engineering equivalent, and hand microassembly, while feasible, is probably not suitable for large numbers. Efficient assembly using a miniature transfer line may be possible, but such systems are usually designed for very high outputs. The situation may change if current mm-scale robot projects are successful.*

Construction problems of such small rovers need consideration. At a cm size the rovers overlap well with the scale of current watch movements, and also with a wide variety of miniature sensors. This suggests the techniques for building them are already in existence, or could be devised with minimal modification of existing processes. A combination of MEMS and conventional small scale engineering could be envisaged. Generally autonomous mobile robots in the literature are 1 cm$^3$ or larger, and the constraining effect of commercially available small parts has been noted by many (e.g. [58]).

For smaller, mm scale rovers, things become more difficult. The types of rover envisaged seem likely to require multiple mechanical systems acting orthogonally. These are difficult to form using an essentially planar lithographic processing. One possible solution which is to create planar foldable, or self folding structures (e.g. [156]). While manual assembly is feasible, it would be tedious for large rover numbers, and complications arise with adverse scaling of adhesive forces. For large volume manufacturing of mm and sub-mm robotic systems, a combination of lithography and a miniature transfer line might allow efficient production, but it is uncertain whether the numbers of mm scale microrovers would be high enough to justify this. (A transfer line is a production line of automated assembly and machining devices which are linked by automated product transfer devices). This sort of equipment is used for wire bonding mm scale integrated circuit chips – and operates on a similar size scale to the assembly of mm-scale rovers. However throughputs are typically very high in order to recoup high capital costs.

The 1980s MIT group suggested building small subsumption-style robots dedicated to assembly [5]. The first generation would be hand assembled, with subsequent generations assembled by previous generations. The design effort would be larger than for a transfer line, as mechanisms for dealing with random component orientation, and to hold the two components to be joined rigidly would be needed.

A detailed study to assess the minimum number of microrobots needed to assemble themselves and the microrovers would be required.

Current work on mm scale autonomous robots is limited, due to the afore mentioned fabrication difficulties, with many working on solving non-scale dependent problems using larger experimental robots. 5 mm, 10 mg, solar powered jumping robots have been investigated by [157]. These used elastic energy storage and MEMS electrostatic motors with friction clutches. A 4x4x7 mm jumping robot, with a limited number of jumps, has been developed using porous Si/ClO$_4^-$ microfabricated fuel [158]. These and [159]'s 10 mg robot have utilised a planar design, so tend to be long and wide for their weight. Several I-swarm prototypes have been built to assess manufacturing problems. The size is 4x4x3 mm, and includes a solar cell, capacitors, control system, IR communications and a crude manipulator [160]. The eventual target is a mass producible autonomous 2x2x1 mm robot [161]. There is also a major Army Research Laboratory program to develop a mm scale, 10 mg autonomous flying robot for reconnaissance use. It will fly using flapping piezoelectrically driven wings and be powered by thin film lithium batteries. The power electronics and control microelectronics will utilise thinned chips [112, 162]. If these programs are successful, they are likely to generate a variety of technologies which could be utilised by mm scale planetary rovers.

*xi) The terrestrial receivers*

*Any planetary exploration campaign that aims to simultaneously explore multiple different targets throughout the solar system will require expansion of deep space communications infrastructure. This can be done relatively cheaply with arrays of small, relatively low cost dishes.*

One problem that may arise with a successful small orbiter-miniature rover program, is the simultaneous exploration of multiple moons, planets and asteroids overwhelming current deep space networks. The present network was constructed when the state of the art receiver was a large dish, with examples up to 70 m in diameter. Costs rapidly escalate with dish diameter (and escalate more rapidly than surface area). However modern electronics allows the construction of dish arrays with equivalent resolving and collecting power using smaller dishes. One concept is to use a 12 m dish as an array element. 7 such dishes would replace one 30 m dish at a cost of about $10.5 million [163]. This suggests that expansion of the number of terrestrial receivers should not be a problem.

**Conclusion**

*A rover size of 10-20 mm has a high degree of scientific utility. APX and natural γ based elemental compositions are obtainable in reasonable time periods. Raman and infrared spectroscopy will give mineralogical/molecular compositions. Sufficient power is available to abrade away weathering patinas, and supply power to various instruments. At this size no novel manufacturing technology is required. Below this size compositional analysis becomes problematic, and communications becomes increasingly difficult, and at mm scale, transfer of sufficient data to orbit becomes a limiting factor. Mm scale rovers can still effectively gather image data, but mass manufacture may require extensive technology development. Given cm scale microrovers and 4-20 kg scale orbiters exploration of multiple surface points on the majority of large bodies in the solar system is potentially feasible on a total mass budget on the ten tonne scale.*

| Sensor/system Scale | Limit | | Density | Scaling |
|---|---|---|---|---|
| **UV** | 30 µm | Diffraction | | L |
| **Optical imaging** | 0.175mm | Diffraction | | L |
| **<3 µm IR imaging** | 0.75 mm | Diffraction | | L |
| **<3 µm IR spectroscopy** | 2.4 mm | Diffraction | 0.33 | L |
| **Raman** | 8 mm | Diffraction (power) | 0.33 | L |
| **Natural γ** | 1cm | | 7.4 ($Tl_6SeI_4$) 7.2 (BGO) | $L^{>3}$, <1cm $L^3$ 2-15cm $L^2$ >15cm |
| | | 24 hr analysis. Photoelectron escape[1] | | |
| **APX[2]** | 2-4 mm | 1 hr analysis, shielding | | $L^2$ |
| **10-20 µm IR spectroscopy** | 8-16 mm | Diffraction (*laser fabrication, power*) | 0.33 | L, (*3 mm QCL*) |
| **Abrasion** | 2 mm | Power (1 AU value) | | |
| **RTG** | 20 mm | Thermal gradient | 2 | $L^3$ (higher at small size) |
| **RTG vacuum** | 2-4 mm | $K_\alpha$ Shielding | | |
| **Betavoltaic** | 1 µm | 5.7 keV β absorption | | |
| **Communication** | 1 mm | Receiver thermal expansion | | $L^5$ (RTG), $L^4$ (solar) |
| **Computation** | 0.2 mm | integration (2d)/power delay product | | |

*Table 7.*

[1] 0.5 cm $HgI_2$ photoelectron escape limit for sub 4.5 MeV γ rays.
[2] APX Z~ B onwards, limited by atmospheric gases

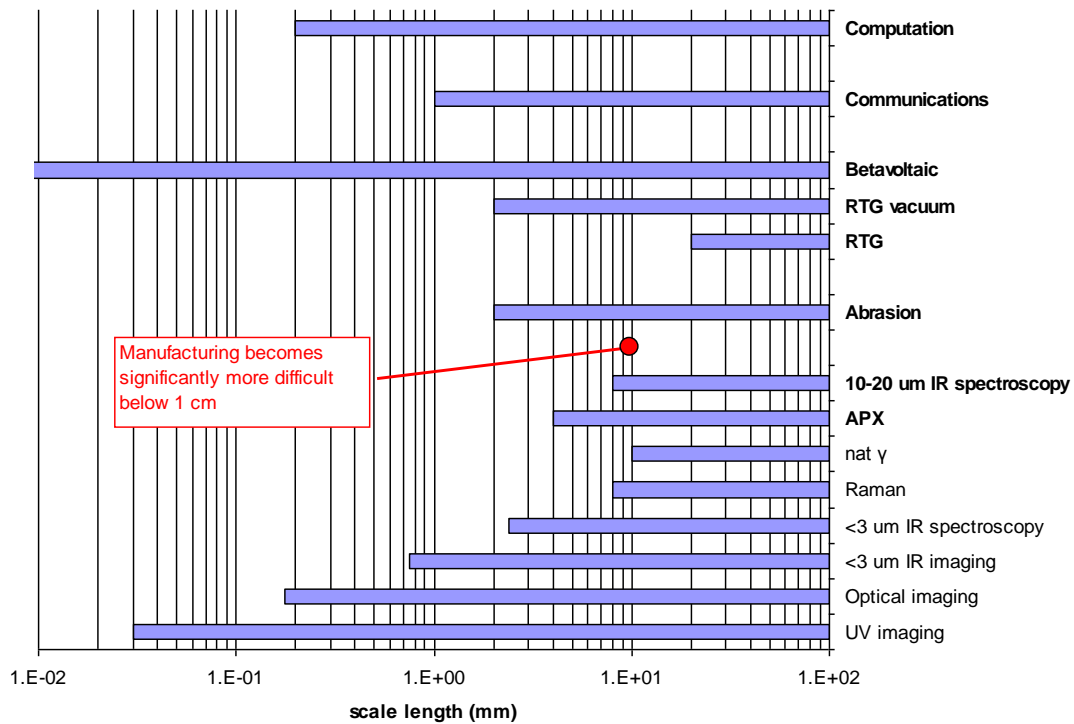

*Figure 14 shows the minimum size at which various rover technologies are feasible, and table 7 the limiting factors. Minimum sizes are not shown for solar power and batteries as these can be scaled down to micrometer size. The minimum rover size is set by a communications limit to about 1 mm, and this will be further limited by solar flux or battery temperature, restricting such "gnat sized" rovers to the inner solar system – at least until more advanced betavoltaics are available. In addition they would be only able to send imaging data. A 1 cm size allows inclusion of a wide range of composition determining tools, as well as requiring a much lower light intensity to power data transmission, extending solar powered rover operation into the outer solar system. At the minimum current RTG size, rovers are capable of performing all the tasks examined.*

*Microfluidic systems have not been plotted, as size is dependent on complexity of analyses, but a minimal channel size will be on the order of 10 μm, suggesting a possible minimum microreactor size of 100x100x10 μm. The major limiting factors will be solvent vaporisation/freezing, and handling and representativeness of very small samples.*

So how small can we make a rover using existing technology? A scale size of about 1-2 cm will allow imaging, composition determination using APX, infrared and/or Raman spectroscopy, and γ spectroscopy combined with sufficient power for surface abrasion and transmission of >100 bits/s to orbit – more than the minimum needed for exploration of a scientifically novel environment. Construction should be relatively straight forward, using at most only minor modifications of existing techniques. Point to point travel with autonomous navigation and manipulator operation is feasible.

It is important to know how mass scales with characteristic length, and this is likely to vary by a factor of 10-100, from a design which unfolds solar panels, dish antennae and effectors (here assumed to unfold, for simplicity, to a size of ~ L), to a rover specialised for high energy gamma ray spectroscopy, where most of the volume is occupied by a high Z detector as dense as 7.4 gcm$^{-3}$ (e.g.Tl$_6$SeI$_4$).

Modest developments in battery technology should allow survival of freeze-thaw cycles, though this is not entirely necessary, as the use of capacitors or simple non-operation during dark periods is an option.

Improvements in wide band gap semiconductors, both in terms of temperature tolerance, integration density and power dissipation are likely, which will allow ambient Venus surface and Mercury equatorial temperature miniature rovers, as well as larger landers and rovers using electronics with >10$^6$ devices per cm$^2$ and operating at <0.1 W. Scintillators, or high (>3.3 eV) band gap semiconductors will be needed for energy dispersive sensors which can operate at up to 500$^o$C.

Rovers smaller than 5-10 mm will have problems utilising Raman/IR spectroscopy, and cleaning surfaces to establish compositions. Mm scale tunable IR sources are possibility, though QCLs appear size limted to >3mm, and >10 μW power. APX size may be limited to larger than 3-4 mm by the need for source K$_α$ shielding. γ ray detector size will be too low to allow sufficient count rates, and to absorb high energy photoelectrons. Difficulties arise with scaling of conventional RTGs. Betavoltaics require considerable development. In the outer solar system, solar photovoltaics will provide insufficient power to maintain high data rates. Manufacturing will also be a problem, as discussed earlier. Such rovers are, at present, for use in special situations, or are best reserved for the future, when microassembly of multiple microelectromechanical systems is better developed. Scaling limits are summarised in table 7 and Fig.14.

Interestingly several mobile rover scale categories emerge:

1) Sub mm - imaging only, and communications to a nearby lander.

2) 1 mm to 10 mm - imaging in the inner solar system, with at 2.4 to 4 mm the determination of elemental composition (from B up) using APX and some mineralogical/molecular compositional information from shortwave IR spectroscopy. QCLs possible as tunable IR source at > 3mm.

3) 10mm to 100mm - imaging and determination elemental and mineralogical/molecular compositions throughout the solar system. Longer wavelength (10-20 μm) IR spectroscopy, Raman spectroscopy and natural gamma spectroscopy become feasible. *Construction becomes much easier due to the availability of components and manufacturing experience at this scale.*

4) >100 mm allows the addition of mass spectrometers and LIBS (see Fig. 14 and Table 7).

Areas for future work include <100 mm terrestrial prototypes to assess communications and locomotion strategies, and cryogenic/high temperature prototypes to assess extreme temperature performance of basic mechanical and energy storage systems. Additionally, small lander prototypes for testing miniaturised rockets for landing in a vacuum environment are needed. Development of III/V or SiC ICs to support at least subsumption-style rover complexity, as well as high temperature optical and energy dispersive sensors will be vital for ambient temperature Venus surface missions at any size scale.

At an orbiter mass of 4 kg and a scale size of 1 m, communications with rovers on the surface and with Earth is not a problem, though radioisotope powered orbiters will provide significant mass savings beyond the orbit of Saturn. Earth-end communications should probably not be a problem, and it may be possible to envisage 50 or more orbiters scattered throughout the solar system relaying data from 5 or more times as many rovers at any one time.

Assuming that most of these rovers operate in the outer solar system, and that the average chemical delta V requirement is 12-14 kms$^{-1}$ [164], 50 rovers per planet or large moon (10 for Venus) a mass of perhaps 10,000 kg delivered to low Earth orbit may suffice. It is hoped that by the time such a program is underway low cost orbital access to allow cheap high mass interplanetary missions will have been achieved, or the scientific questions raised by a low mass exploratory program will spur the development of such technology.

**Acknowledgments**

I would like to thank the following who have assisted with various aspects of the paper, Prof Ray Burgess, Prof. Jamie Gilmour, Dr Martin Guitreau, Dr Katherine Joy, Dr Peter Roberts and Dr Robin Sloan.